\documentclass{SCIS2023}
   
\begin{document}
\ArticleType{RESEARCH PAPER}

\Year{2022}
\Month{}
\Vol{}
\No{}
\DOI{}
\ArtNo{}
\ReceiveDate{}
\ReviseDate{}
\AcceptDate{}
\OnlineDate{}


\title{Ensemble Successor Representations for Task Generalization in Offline-to-Online Reinforcement Learning}{}


\author[1]{Changhong Wang}{}
\author[1]{Xudong Yu}
{hit20byu@gmail.com}
\author[2,4]{Chenjia Bai}{}
\author[2]{Qiaosheng Zhang}{}
\author[3]{Zhen Wang}{zhenwang0@gmail.com}

\AuthorMark{Author A}

\AuthorCitation{Author A, Author B, Author C, et al}


\address[1]{Space Control and Inertial Technology Research Center, Harbin Institute of Technology, Harbin {\rm 150001}, China}
\address[2]{Shanghai Artificial Intelligence Laboratory, Shanghai {\rm 200232}, China}
\address[3]{Northwestern Polytechnical University, Xi’an 710072, China}
\address[4]{Shenzhen Research Institute of Northwestern Polytechnical University, Shenzhen 518057, China}

\abstract{In Reinforcement Learning (RL), training a policy from scratch with online experiences can be inefficient because of the difficulties in exploration. Recently, offline RL provides a promising solution by giving an initialized offline policy, which can be refined through online interactions. However, existing approaches primarily perform offline and online learning in the same task, without considering the task generalization problem in offline-to-online adaptation. In real-world applications, it is common that we only have an offline dataset from a specific task while aiming for fast online-adaptation for several tasks. To address this problem, our work builds upon the investigation of successor representations for task generalization in online RL and extends the framework to incorporate offline-to-online learning. We demonstrate that the conventional paradigm using successor features cannot effectively utilize offline data and improve the performance for the new task by online fine-tuning. To mitigate this, we introduce a novel methodology that leverages offline data to acquire an ensemble of successor representations and subsequently constructs ensemble Q functions. This approach enables robust representation learning from datasets with different coverage and facilitates fast adaption of Q functions towards new tasks during the online fine-tuning phase. Extensive empirical evaluations provide compelling evidence showcasing the superior performance of our method in generalizing to diverse or even unseen tasks.}

\keywords{Offline Reinforcement Learning, Online Fine-tuning, Task Generalization, Successor Representations, Ensembles}

\maketitle

\section{Introduction}

Reinforcement Learning (RL) has emerged as a powerful approach for tackling complex sequential decision-making problems in various domains, such as games \cite{alphago2017}, robotics \cite{scis2022}, manipulation \cite{scis2022manipulation}, and autonomous driving \cite{autodriving, scis2023}. However, most applications rely on extensive online interactions with the real environment or high-fidelity simulators, which can be infeasible or cost-expensive in real-world scenarios. Recently, offline RL \cite{offlinesurvey} provides a promising solution to address this problem by learning an offline policy from a fixed dataset, without requiring online interactions. After that, the learned offline policy provides a good initialization for subsequent online fine-tuning. This paradigm allows efficient utilization of pre-collected data to provide an initialized policy and further improves the policy via limited online interactions \cite{AWAC2020, off2onrl2022}. 

Despite the potential of the offline-to-online paradigm to improve the sample efficiency in decision-making problems, existing approaches \cite{off2onrl2022,iql2022, pex2023, calql2023} are limited to the setting  where the tasks of offline pre-training and online fine-tuning remain the same. In real-world applications, it is common to possess an offline dataset from a specific task while desiring policy generalization across various tasks. In such scenarios, the main challenge lies in effectively leveraging the information embedded within the offline datasets to benefit new tasks. For model-free methods, the policies and value functions are typically learned for a specific task, thus fine-tuning them may cause poor performance on the new task due to overfitting on the offline datasets. For model-based methods, while their transition functions are invariant for different tasks, they necessitate accurate models for planning in downstream tasks and an exploratory dataset that covers the entire state-action space. Consequently, new approaches are required to facilitate generalization to new tasks during the fine-tuning stage while making full use of offline data.

Our study focuses on the task-generalization problem in an offline-to-online setting. In our work, we first explore the successor representation \cite{dsr2016, SRgpi2017}, which is a reward-agnostic representation that implicitly captures the underlying dynamics to predict future outcomes. By decoupling the dynamics from the reward function, successor representations enable rapid adaptation of the value function to novel tasks. Despite this advantage, we reveal that vanilla methods based on successor representation cannot learn effective policies during the fine-tuning phase. Moreover, we observe that the coverage of the offline data significantly impacts performance in offline-to-online generalization. Specifically, representations learned from the offline data with a narrow distribution is hard to generalize to novel tasks. Therefore, it is important to design an algorithm capable of conducting policy adaptation in new tasks with limited online interactions, while being robust to the data distributions of offline datasets.

To this end, we propose a novel approach that combines Ensemble networks with Successor Representations to perform Offline-to-Online adaptation (ESR-O2O) in various downstream tasks.
ESR-O2O adopts ensemble architecture to enhance the diversity of successor representations and value functions. This alleviates the dependency on data coverage or behavior policies during the offline training stage, allowing for learning useful and transferable representations even in scenarios where the available offline data exhibits a narrow distribution. During the online-adaptation stage, we keep the successor representations fixed and update task-specific parameters to learn value functions and policies for the downstream tasks. Theoretically, we establish that the optimality gap during online fine-tuning is bounded. Through empirical evaluations and comparisons, we show that ESR-O2O significantly outperforms existing RL methods for task generalization in offline-to-online settings. 

Overall, our contributions are threefold. 1) To the best of our knowledge, our work is the first to investigate the task generalization problem in the context of offline-to-online RL, specifically focusing on pre-trained agents derived from a single offline environment. By addressing the reward gap, our work provides valuable insights into bridging the gap between offline pre-training and generalization in RL.
2) We introduce a novel approach that leverages ensembles to learn successor representations from datasets, thereby enhancing the robustness of the learned representations, especially when confronted with narrow offline distributions. This ensemble-based framework mitigates the dependence on data coverage and enhances the adaptability across various RL tasks, leading to improved performance in online fine-tuning. 3) We provide a substantial body of empirical results and a theoretical analysis to validate the feasibility and effectiveness of our method. We demonstrate that ESR-O2O effectively handles reward changes during fine-tuning, surpassing alternative approaches in terms of performance and sample efficiency.

\section{Preliminaries}
\subsection{Reinforcement Learning}

We adopt the episodic Markov Decision Process (MDP) framework to formulate the sequential decision-making problem \cite{wang2022modelling, chu2022formal}. Specifically, we define $M=(\mathcal{S}, \mathcal{A}, R, \mathcal{P}, \gamma, T)$, where $\mathcal{S}$ and $\mathcal{A}$ represent the state and action spaces, respectively. The transition dynamics of the environment is captured by the function $\mathcal{P}$, while $R$ denotes the reward function. The discount factor $\gamma$ represents the agent's preference for immediate rewards versus future rewards, and $T$ specifies the length of an episode.

In online RL \cite{scis2019}, the agent first observes the current state $s \in \mathcal{S}$ of the environment and applies an action $a \in \mathcal{A}$ to the environment based on its policy $\pi(a|s)$. The environment then gives the next state based on $\mathcal{P}$ and provides the agent with a scalar reward $r$. The agent repeats this process over time, with the goal of maximizing the expected cumulative reward, defined as $\mathbb{E}[\sum_{t=0}^{T}\gamma^t R(s_t,a_t)]$.
To quantify the value of a policy $\pi$, we define the state-action value function as follows:
\begin{equation}
    Q^\pi(s,a) = \mathbb{E}\left [\sum_{t=0}^{T} \gamma^t R(s_t, a_t)|s_{t+1}\sim \mathcal{P}(\cdot|s_t,a_t), a_t\sim \pi(\cdot|s_t)\right ].
\end{equation}
This function represents the expected cumulative reward starting from state $s$ and taking action $a$ under policy $\pi$. It can be updated by minimizing the temporal difference (TD) error $\mathbb{E}[(Q-\mathcal{T}Q)^2]$, where $\mathcal{T}Q=r+\gamma\mathbb{E}[\max_{a'}\bar{Q}(s',a')]$ is the Bellman operator and $\bar{Q}$ indicates the target network.

In offline RL, given a batch dataset $D$ consisting of tuples $(s,a, r,s')$, where $r\sim R(\cdot|s,a)$ and $s'\sim \mathcal{P}(\cdot | s,a)$, agents are expected to find a policy $\hat{\pi}$ based on $D$ so as to minimize the expected sub-optimality with respect to the optimal policy $\pi^*$, i.e., $\mathbb{E}_D[J(\pi^*) -J(\hat{\pi})]$, where the expectation is taken with respect to the randomness in the dataset. However, the bootstrapping process leads to an overestimation of the value for out-of-distribution (OOD) actions, because the bootstrapped error cannot be corrected without online interactions. Due to this, off-policy algorithms fail to learn useful policies from the static dataset. To overcome the bootstrapped error, several offline RL methods apply policy constraints \cite{bcq2019, brac2019, td3bc2021} or conservative regularization to values \cite{cql2020}, but may cause over-conservative estimation. There is another line of works \cite{edac2022, msg2022, sacbcN2023, rlpd2023} that utilize ensembles for value functions to capture the epistemic uncertainty and attain favorable performance in many tasks. Specifically, the TD error in these methods becomes $\mathbb{E}[(\mathcal{T}Q - Q_i(s,a))^2]$, where the TD target could be shared for each ensemble network \cite{edac2022}, i.e. $\mathcal{T}Q=r+\gamma\min \bar{Q}_i(s,a)$ or independently learned \cite{msg2022}, i.e. $\mathcal{T}Q_i=r + \bar{Q}_i(s,a)$.

\subsection{Successor Representation}\label{sec:sr}
As an appealing approach to task transfer, successor representations (SR) \cite{sr1993}, especially successor features (SF) \cite{dsr2016} for continuous state space, have been proposed as a generalization for the value function. Let $M(s_t,a_t,s')=\sum_{i=0}^\infty \gamma^i p(s_{t+i}=s'|s_t,a_t)$ be the successor representation, defined as the discounted occupancy of state $s'$, averaged over trajectories initiated in state $s$, then the state-action value function can be expressed as:
\begin{equation}
    Q(s,a)=\sum_{s'}M(s,a, s')R(s').
\end{equation}

The SR allows the decoupling of the dynamics of an MDP from its reward functions.
The SR can intuitively be thought of as a predictive map that encodes each state in terms of the other states that will be visited in the near future. The SR can be learned in a similar way to temporal difference learning:
\begin{equation}
    \delta_t(s')=\mathbbm{1}[s_t=s']+\gamma M(s_{t+1},s')-M(s_t,s'),
\end{equation}
where the error is the discrepancy between observed and expected state occupancy.
The expected occupancy for states that are visited more frequently than expected should be increased, whereas the expected occupancy for states that are visited less frequently than expected should be decreased.

The successor feature (SF) extends SR to handle continuous state spaces by assuming that the reward function can be expressed as a linear combination of features $\phi$ and a weight vector $w$:
\begin{equation}
    r(s_t, a_t, s_{t+1}) = \phi(s_t,a_t,s_{t+1})^{T}w.
    \label{eq:reward}
\end{equation}
Given the above decomposition, the Q-function can be rewritten by:
\begin{equation}
    Q(s,a)=\mathbb{E}\left[\sum_{i=0}^\infty \gamma^i r(s_i,a_i)\right]=\mathbb{E}\left[\sum_{i=0}^\infty \gamma^i \phi^{T} w \right]=\psi(s,a)^{T}w.
\end{equation}

In practice, the features $\psi$ can be learned in many ways such as regression, using transition models or auto-encoders. The successor features $\psi$ can also be learned by Bellman backups:
\begin{equation}
    \bar{\psi}(s,a) = \phi(s,a) + \gamma \psi(s_{t+1}, a'), \quad a'\sim \pi(\cdot|s_{t+1}).
\end{equation}

The above formulation provides the possibility of quickly evaluating a policy $\pi$, thus being a promising way to handle the reward gap problem. When the reward function changes, we can keep the learned successor representation and re-learn the weight vector $w$ by regression to obtain a new value function. However, the linear decomposition of the reward function is limited.
Prior work addresses the reward generalization problem by considering a linear decomposition of the reward function and the value function. This modeling is restrictive and may fall short in capturing the complexity of the changed environment.

\subsection{Problem Setting}
In this paper, we focus on the offline-to-online setting, where an agent is first pre-trained with offline datasets and then fine-tuned by interacting with the environment. Several works \cite{AWAC2020, JSRL2022} found that the agent exhibits a huge dip at the initial stage of online fine-tuning, which is not expected in real applications. More importantly, we put attention to the \emph{reward generalization} problem that the pre-trained agents are fine-tuned in a new environment with the same dynamics but different reward functions.

Here, we present a clear formulation for our problem setting. Let $M_i=(\mathcal{S},\mathcal{A}, \mathcal{P}, R_i, \gamma)$ be the MDP in the offline training stage and logged data are collected following the reward function $R_i$. Then we introduce a different MDP as $M_j=(\mathcal{S},\mathcal{A}, \mathcal{P}, R_j, \gamma)$ for online fine-tuning of the policy after pertaining in $M_i$. We remark that $M_i$ and $M_j$ are different tasks with the same dynamics but different reward functions. In the offline training stage, agents learn representations, value functions, and policies from offline data. To perform policy generalization
in the fine-tuning stage with a different task $M_j$, a robust and favorable performance is expected to provide a good initialization and fast adaptation. 

\section{Related Work}

\paragraph{Offline-to-Online RL.} Our work contributes to offline-to-online RL by fine-tuning pre-trained offline agents through online interactions. Previous studies have primarily addressed distributional shift, which refers to the disparity between offline data and online transitions.
Various approaches have been proposed to mitigate the performance drop during the initial fine-tuning stage caused by distributional shift. These approaches include adaptive adjustment of behavior cloning weights \cite{adaptiveBC2022}, reconstruction of the replay buffer and sampling methods \cite{off2onrl2022}, modification of the update target of the critic network \cite{AWAC2020,iql2022, calql2023}, and adaptive policy composition \cite{pex2023}. Recently, E2O \cite{e2o2023} and PROTO \cite{proto2023} have achieved significant performance by integrating offline pessimism and online optimism with ensembles, or using an iteratively evolving regularization term and performing a trust-region-style update. While both E2O and our approach utilize ensembles, their goals and solved problems differ. E2O employs Q ensembles to perform pessimism to prevent over-estimation during offline training and to encourage exploration during online fine-tuning. However, our approach use ensembles to model diverse patterns of successor representations and Q functions, allowing robust representation learning despite offline data limitations. More importantly, these offline-to-online algorithms mainly address scenarios with consistent offline and online environments. The issue of distributional shift and generalization across different tasks has received limited attention. To the best of our knowledge, our work is the first to investigate the problem of reward generalization in the offline-to-online setting.

\paragraph{Task Generalization.} Our work aims to deal with the task generalization problem. While recent studies \cite{augWM2021, vgdf, rewardinformer2023, gsf2022} have made progress in addressing generalization and distributional shift challenges, they focus on specific aspects of the offline-to-online learning process. These works either consider minor changes in dynamics \cite{augWM2021}, emphasize generalization in an online setting \cite{rewardinformer2023}, or primarily focuses offline learning \cite{gsf2022}, which is constrained by dataset quality. On the other hand, successor features are utilized to handle learning in dissimilar environments \cite{SRnavigation2017, sfde2021}. But these approaches consider this problem in an online setting and assume the availability of an effective exploration policy during the learning process. In contrast, our work focuses on applying pre-trained agents, including learned representations, value functions, and policies, to a different online environment. Furthermore, many existing methods that utilize successor representations also assume the use of a set of training tasks, while our work does not impose such a limitation. By focusing on the reward generalization problem and the offline-to-online setting, our work aims to bridge the gap between offline pre-training and online fine-tuning, contributing to a deeper understanding of generalization in reinforcement learning.

\paragraph{Successor Representation.} Our method leverages successor representation to capture the environmental dynamics, enabling generalization across different reward functions. Successor representations serve as predictive representations that summarize the successive features to follow. They also provide mechanic explanations similar to human understanding \cite{SRhuman2017}. Several previous studies have combined successor representation with Generalized Policy Improvement \cite{SRgpi2017} to transfer behaviors across navigation tasks \cite{SRnavigation2017}.  Recently, the unsupervised representation learning paradigm combined with the successor measure is discussed \cite{fb2021, fb2023}, avoiding learning basic features for conventional successor features. Nevertheless, these works often rely on effective exploration policies or exclusively diverse datasets. In contrast, our work analyze learning representations from various types of dataset, expanding the applicability and potential of successor representations.

\paragraph{Ensemble.} Our work is also related to ensemble-based methods. In model-free RL, ensemble methods have gained considerable interest for estimating epistemic uncertainty for action-value estimates. In online-RL, ensembles are frequently employed to enhance exploration \cite{bootstrappedDQN2016, ucbexploration2017} by encouraging agents to seek out actions with the highest variance in estimated values. This is achieved by constructing a distribution of action-value estimates using the ensemble and acting optimistically with respect to the upper bound \cite{redq2021,bai2021principled,bai2021dynamic,CUCB}. In offline-RL, ensembles of Q functions or environmental models are used to obtain conservation estimation from the dataset \cite{edac2022, msg2022, sacbcN2023, robustO2O}. Deep ensembles \cite{deepensemble2019} are shown to effectively capture epistemic uncertainty arising from incomplete information and approximate the true posterior distribution. By leveraging ensembles, the diversity of networks is enhanced, thereby mitigating estimation bias for the value function \cite{maxminQ2020}. Inspired by this, our work adopts ensemble networks to improve the estimation of successor representations and value functions.

\section{Motivating Examples}\label{sec:motivating}

In this section, we provide motivating examples to illustrate the challenges in offline-to-online RL, especially when dealing with reward gaps. We also highlight the limitations of vanilla successor representation based methods when learning from offline data. First, we use a tabular case to show that existing offline-to-online RL approaches struggle with reward generalization. In contrast, successor representations can enhance adaptive value function learning. Furthermore, we explore the limitation of vanilla SR based methods in task generalization after pre-training on a single offline environment. Specifically, we find that the coverage of the offline data affects both  pre-trained performance and online fine-tuning performance. 

\begin{figure}[t]
\centering
\begin{minipage}[c]{\textwidth}
\centering
\includegraphics[width=0.9\linewidth]{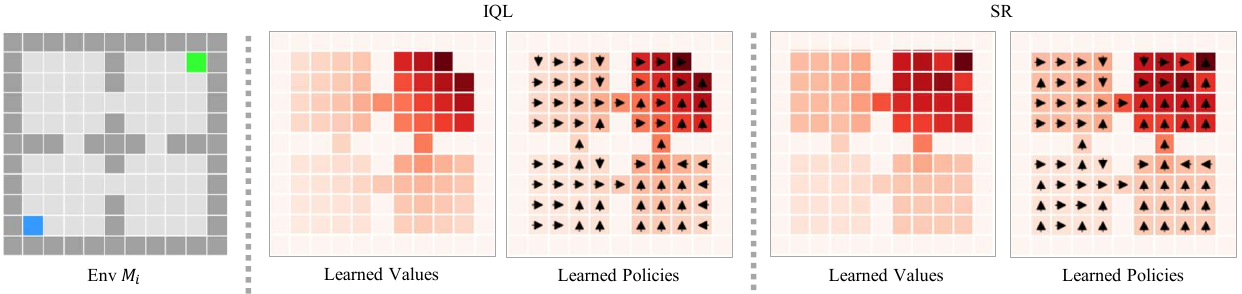}
\caption{Offline pre-training in a four-room maze navigation task. In the left picture, the blue block indicates the starting point, and the green one indicates the goal. Two middle pictures describe the values and policies learned by IQL from offline data, where the arrow represents the action to be taken at each position. The two right pictures show the values and policies learned by vanilla SR based methods. After offline training, both IQL and SR-based methods can learn correct values and the optimal policy.}
\label{fig:offline_env}
\end{minipage}

\begin{minipage}[c]{\textwidth}
\centering
\includegraphics[width=0.9\linewidth]{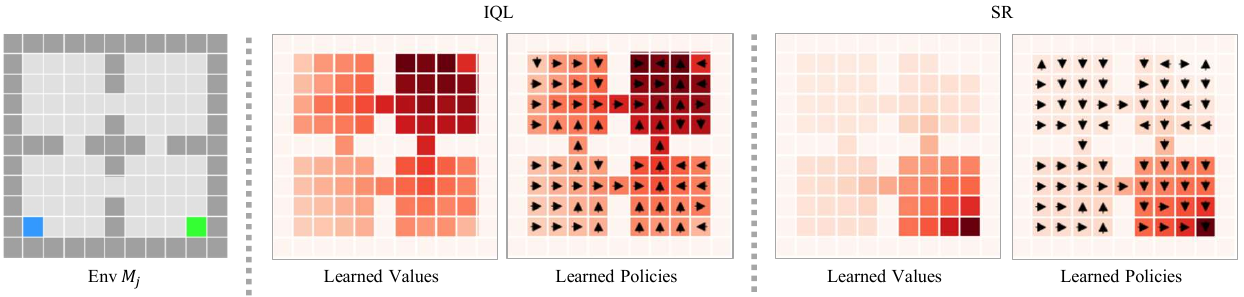}
\caption{Online fine-tuning in a navigation task with a different goal. When the goal changes, the reward function changes accordingly. In this setting, IQL fails to learn correct values and policies in the new environment, while SR based methods re-learns of the values and leads to effective policies.}
\label{fig:online_env}
\end{minipage}
\end{figure}

\paragraph{Grid World Example.} 
Consider a navigation task in a four-room environment, with each room divided into a $4\times 4$ grid of cells, as depicted in Figures \ref{fig:offline_env} and \ref{fig:online_env}. In this scenario, the blue block marks the starting point, and the green block indicates the goal. Our paper focuses on a specific setting where the agent has access to a dataset from the environment $M_i$. The objective is to apply pre-trained policies from the offline dataset to a new environment $M_j$. In the grid world example, $M_i$ and $M_j$ possess different reward functions, achieved by altering the goal's location while keeping the transition dynamics constant.

To illustrate the challenge of task generalization in offline-to-online RL, we evaluate two methods: the state-of-the-art algorithm IQL and a vanilla successor representation based method. In Figure \ref{fig:online_env}, IQL successfully learns the optimal values in the original environment $M_i$ during the pre-training phase. However, when attempting to fine-tune the learned values in the new environment $M_j$, IQL fails to generate accurate value estimations. This limitation arises from the agent's specialization in training for specific reward functions, hindering its ability to generalize effectively to novel reward functions. In contrast, values learned by the vanilla SR-based method show adaptability to new environments with just a few fine-tuning steps. This comparison highlights the limitations of current offline-to-online RL methods and the potential of SR-based methods in addressing the task generalization problem.

\begin{figure}[t]
    \centering \includegraphics[width=0.5\linewidth]{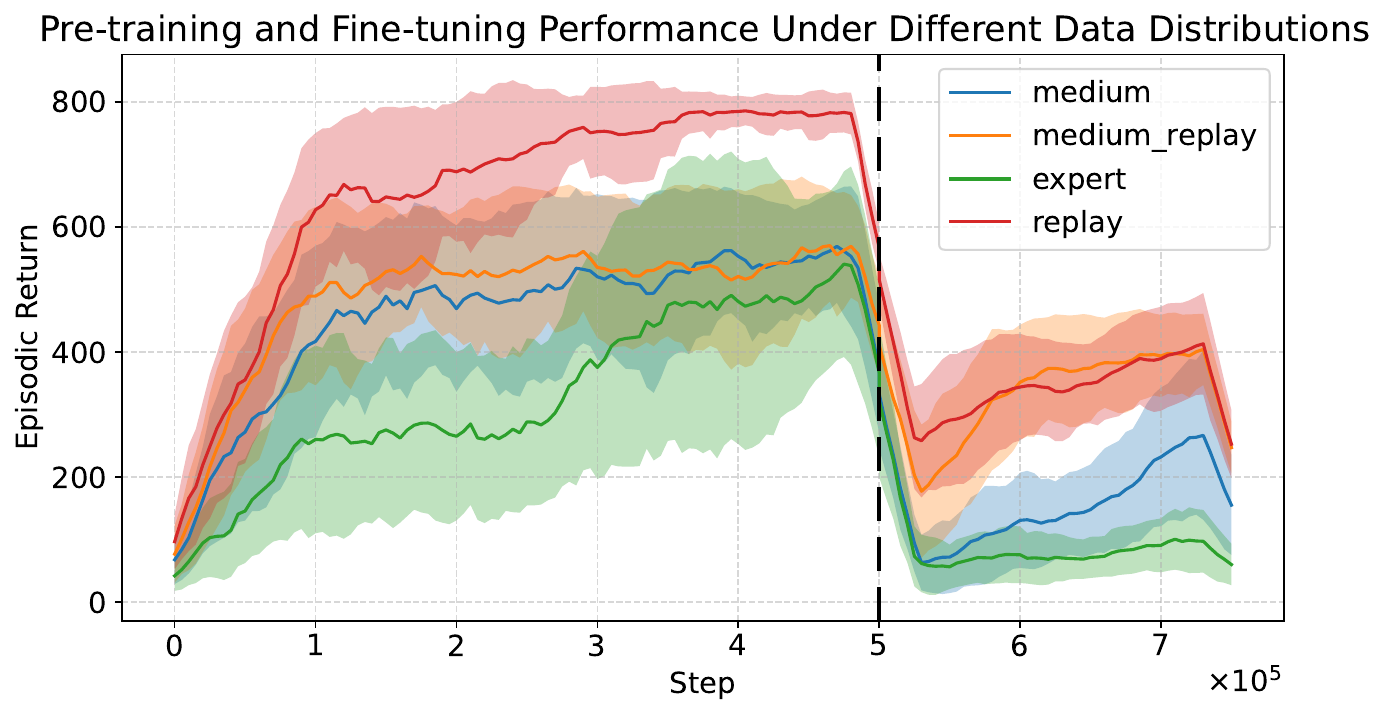}
    \caption{Pre-training and Fine-tuning performance of vanilla SR based methods under different data distributions. The black dashed line indicates the transition from offline pre-training to online fine-tuning when the task changes. The shaded area represent the variance of performance across multiple experiments conducted with 5 random seeds. Representations learned from more diverse datasets, such as `medium-replay' and `replay' data, exhibit superior performance compared to representations learned from narrower datasets (`medium', `expert').}
    \label{fig:motivating}
\end{figure}

\paragraph{The Impact of Offline Data.} Despite the advantage of adapt to new tasks, challenges persist in learning from offline data using vanilla SR based methods (as presented in Section \ref{sec:sr}). Here, we consider a more complex scenario with continuous state and action space, where representations and values are expressed by neural networks. Specifically, we test a quadruped robot's ability to fine-tune its performance from a rolling task to a walking task after training on an offline dataset. This transition introduces a reward gap, presenting an offline-to-online RL challenge. To investigate this, we use four types of offline datasets with varying coverage: `medium' data from a medium-level agent, `medium-replay' data encompassing all experiences in training a medium-level agent, `expert' data generated by an expert-level agent, and `replay' data including all experiences in training a expert-level agent. Our approach involves the initial extraction of successor representations from the offline data, followed by policy and value function fine-tuning based on online interactions.

Address value overestimation during offline pre-training due to distributional shift is crucial. In vanilla SR based methods, both representations and values are updated via bootstrapping. The bootstrapping process can result in overestimation \cite{bcq2019}, since the bootstrapped error cannot be corrected during offline learning. To mitigate this, we make two key modifications. We set the minimum of two critic estimates as the TD target \cite{td32018} and employ Layer Normalization to prevent catastrophic overestimation \cite{rlpd2023}. As shown in Figure \ref{fig:motivating}, these improvements help the vanilla SR based method to learn policies from offline data. During the pre-training stage, we observe that the performance is affected by the coverage and quality of offline dataset. For example, agents trained on `replay' data tend to exhibit better performance compared to those trained on `expert' data. 

In addition to the overestimation problem, another challenge is the reward gap, which exacerbates the discrepancy between offline datasets and online transitions. During fine-tuning, we find that agents pre-trained on `replay' or `medium-replay' data outperforms those trained on `expert' or `medium' data. This highlights the significant influence of data coverage on successor representation learning and subsequent fine-tuning performance. In other words, greater diversity in offline data leads to more effective learned representations and improved performance. However, practical concerns arise as there are no guarantees regarding the diversity and quality of offline datasets. Hence, the challenge remains in learning valuable successor representations from datasets with different coverage and quality.

\section{Methodology}

In this section, we present our approaches to address task generalization challenges in offline-to-online RL. First, we analyze the sub-optimality gap during the fine-tuning phase when a reward gap exists. Based on this analysis, we propose a simple yet effective method to improve the ability to generalize to unseen tasks by incorporating ensembles. This method diversifies successor representations and critic networks, reducing the bias in representations and value functions. We then outline the pre-training and fine-tuning process that leverages ensemble successor representations to tackle the task generalization problem.

\subsection{Theoretical Analysis}\label{sec:theo}
To figure out how to handle the task generalization problem, we delve into the theoretical foundations to characterize the sub-optimality gap in the fine-tuning phase. First of all, let us introduce some notations for clarity. Consider the offline environment $M_i$ with reward function $r_i$ and the online environment $M_j$ with $r_j$. We denote $\pi^*_i$ and $\pi^*_j$ the optimal policies in $M_i$ and $M_j$, respectively. We introduce an optimal successor feature $\psi^{\pi^*}$ by considering $Q^{\pi^*}=\psi^{\pi^*}w$. This optimal successor feature represents the future state occupancy by following policy $\pi^*$. In $M_i$ and $M_j$, the optimal successor features can be expressed as $\psi^{\pi^*_i}$ and $\psi^{\pi^*_j}$. Their corresponding optimal value functions are denoted as $Q^{\pi^*_i}_i=\psi^{\pi^*_i}w_i$ and $Q^{\pi^*_j}_j=\psi^{\pi^*_j}w_j$, where $w_i$ and $w_j$ indicate the weight vectors as shown in Equation \ref{eq:reward}.

In our work, we pre-train agents from the offline dataset and then fine-tune them in the online phase. We define the pre-trained successor feature as $\hat{\psi}$ and pre-trained policies as $\hat{\pi}$. Then, we can get the following performance bound during the fine-tuning stage.

\proposition[Sub-optimality Gap.]\label{proposition}{For all $s\in \mathcal{S}, a \in \mathcal{A}$, let the learned value function after the fine-tuning stage be $Q_j^{\pi}=\hat{\psi}w^{\pi}$, then the fine-tuning performance bound can be expressed as:
\begin{equation}
    |Q_j^{\pi^*_j}- Q_j^{\pi}|\leq \|w_j\|_{\infty}\|\psi^{\pi^*_j}-\hat{\psi}\|_{1} + \|\hat{\psi}\|_{\infty}\|w_j-w^{\pi}\|_{1}
    \label{eq:subopt}
\end{equation}
\label{theo:subopt}
}
\proof{The sub-optimality gap can be decomposed according to the triangle inequality and Holder's inequality:
\begin{align}
    |Q_j^{\pi^*_j}- Q_j^{\pi}| 
    & = |Q_j^{\pi^*_j}- Q_j^{\hat{\pi}} + Q_j^{\hat{\pi}} - Q_j^{\pi}| \leq |Q_j^{\pi^*_j}- Q_j^{\hat{\pi}}| + |Q_j^{\hat{\pi}} - Q_j^{\pi}| \\
    & \leq \|w_j\|_{\infty}\|\psi^{\pi^*_j}-\hat{\psi}\|_{1} + \|\hat{\psi}\|_{\infty}\|w_j-w^{\pi}\|_{1}
\end{align}
}

This proposition indicates that the sub-optimality gap in the fine-tuning stage is influenced by the optimality of successor features and the weight vector. The term $\|\psi^{\pi_j^*}-\hat{\psi}\|_{1}$ represents the difference between the optimal successor feature in $M_j$ and the pre-trained successor feature. Since the successor feature captures the dynamics information, this difference could be interpreted as the disparity between the dynamics captured by the offline dataset and the dynamics in $M_j$. The term $\|w_j-w^{\pi}\|_{1}$ quantifies the approximation error for the true reward function in $M_j$. If we retain the pre-trained value function from the offline dataset and assume that $w^{\pi}$ approximate $w_i$ well, then $w_j-w^{\pi}$ contains a transition from $w_i$ to $w_j$. When the reward gap $\|w_i-w_j\|_{1}$ is large, this transition can be challenging due to the gradient propagation mechanism in neural networks. Consequently, the performance of the fine-tuning process can be compromised.

We further bound the two terms in Equation \ref{eq:subopt} under several assumptions. Assuming that the norm of the basic features and the rewards are bounded, i.e. $\|\phi\|_2\leq 1$ and $|r|\leq r_{\max}$, we can establish a similar inference process as in linear MDP. The following lemma demonstrates that the gaps in fine-tuning performance are bounded. Specifically, the reward approximation error depends on the quality and size of the data buffer collected during fine-tuning, while the representation gap is bounded by the estimated error of the dynamics.

\lemma[\cite{psi_lemma2021, w_lemma2011}]{
Let $Z=\sqrt{\lambda}+r_{\max}\cdot \sqrt{2\log \frac{1}{\delta} + d \log (1+\frac{N}{\lambda d})}$, $\Lambda=\lambda I+\sum_{i=1}^{N}\phi(s_i,a_i)\phi(s_i,a_i)^\top$, then the following inequality holds with probability $1-\delta$:
\begin{equation}
    \|w_j - w^\pi\|_{1}\leq \sqrt{d}\|w_j - w^\pi\|_{\Lambda}\leq \sqrt{d}Z,\label{eq:11}
\end{equation}
where $w_j, w^\pi \in\mathbb{R}^d$, $\|w\|_\Lambda=\sqrt{w^\top\Lambda w}$, $\lambda$ is the regularization parameter, and $N$ denotes the size of the data buffer. For successor representations, we have
\begin{equation}
    \|\psi^{\pi_j^*}-\hat{\psi}\|_{1}\leq \frac{\gamma}{(1-\gamma)^2}\|\mathcal{P}^*-\mathcal{P}\|_{1},
\end{equation}
where $\mathcal{P}^*$ and $\mathcal{P}$ denote the true and estimated dynamics, respectively.
}

\proof{For Equation (\ref{eq:11}), we simplify the equation by setting $\bar{w}=w_j - w^\pi$, then we have
\begin{equation}
\begin{aligned}
    \|\bar{w}\|_{1}
    & \leq \sqrt{d}\|\bar{w}\|_2 \leq \sqrt{d} \sqrt{\|\bar{w}\|_2^2+\|\phi^\top \bar{w}\|_2^2}  = \sqrt{d} \sqrt{\bar{w}^\top(I+\phi^\top\phi)\bar{w}} = \sqrt{d}\|\bar{w}\|_{\Lambda} \leq  \sqrt{d} Z.
\end{aligned}
\end{equation}
Please refer to Theorem 16 in \cite{psi_lemma2021} and Theorem 2 in \cite{w_lemma2011} for more detailed proof.}

Based on this lemma, the estimation error of the successor representation and the weight vector is bounded. Since $\|\hat{\psi}\|_{\infty}=\|\sum_{i=0}^{\infty}\gamma^i\phi^\top\|_{\infty}=\|\frac{1}{1-\gamma} \phi^\top\|_{\infty} \leq \frac{1}{1-\gamma}$, the sub-optimality gap in Proposition \ref{proposition} is upper bounded.

\subsection{Randomized Ensembles of Successor Representations and Critic Networks}

In Section \ref{sec:motivating} and Section \ref{sec:theo}, we provide empirical and theoretical insights about learning successor representations from offline data and using them to deal with task generalization in the fine-tune stage. Specifically, we observe that the fine-tuning performance is influenced by the learning process of value functions, represented by weight vectors, and the pre-trained representations. The quality of pre-trained representations depends on the quality and coverage of the offline data, as well as the handling of catastrophic overestimation. To this end, we expect agents to acquire well-performed representations regardless of the offline data distribution and demonstrate robust task generalization capabilities.

In this paper, we propose a novel approach utilizing randomized ensembles of successor representations and critic networks, named by ESR-O2O, to address the challenges discussed earlier. Our design is motivated by several reasons. Firstly, multiple estimates enables characterization for epistemic uncertainty and the lower confidence bound of estimations for representations and values, facilitating efficient and pessimistic learning in the offline stage \cite{pevi2021, pbrl2022}. Secondly, ensemble helps to mitigate the estimation bias \cite{maxminQ2020}, thus reducing the gaps in the sub-optimality bound described in Equation (\ref{eq:subopt}). Thirdly, ensembles are beneficial to enhancing the sample efficiency \cite{redq2021, rlpd2023}. Finally, ensembles impose diversity on the estimation, thereby mitigating the limitation of the coverage of offline datasets.

\begin{figure}[ht]
\centering
\subfloat[Vanilla SR]{\includegraphics[width=0.9\linewidth]{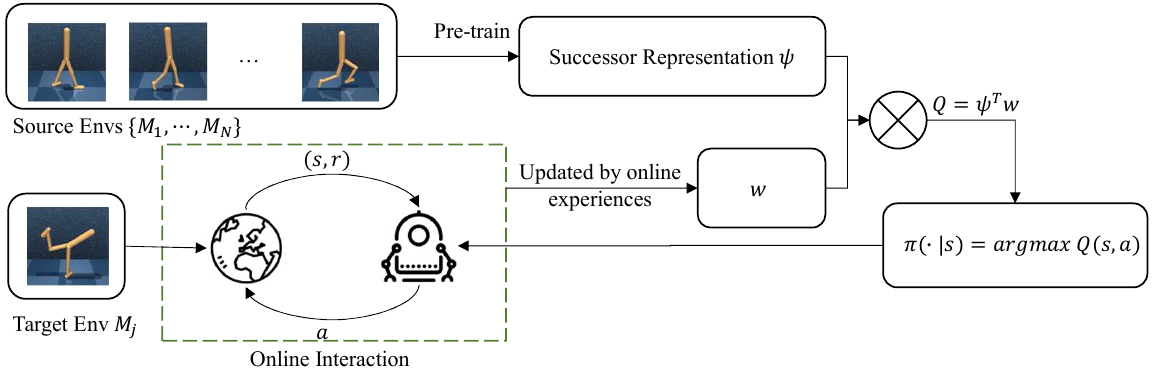}}

\subfloat[ESR-O2O for offline-to-online learning]{\includegraphics[width=0.9\linewidth]{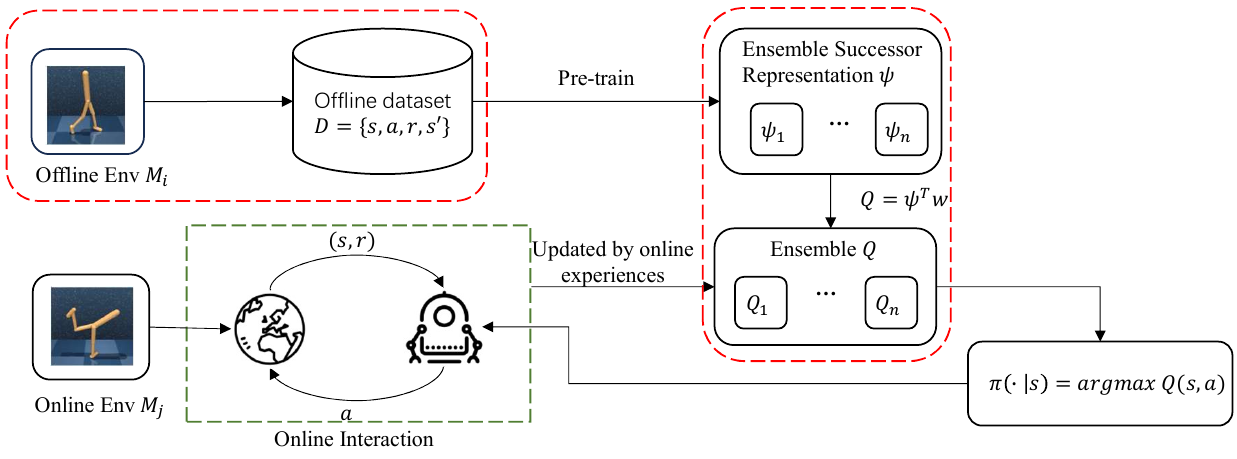}}
\caption{Frameworks of vanilla SR-based methods and ESR-O2O for offline-to-online learning. Red dashed boxes indicate the different parts of ESR-O2O from vanilla SR. While vanilla SR trains multiple representations from source environments, ESR-O2O extract ensemble representations using offline datasets from a single environment. Another difference lies in the construction of Q function, where vanilla SR considers linear composition, and ESR-O2O incorporates ensemble Q functions. Other parts are the same, including fine-tuning the value function with online interactions with $M_j$, which follows a greedy policy $\pi(\cdot|s)=\arg\max Q(s,a)$.}
\label{fig:frame}
\end{figure}

We compare our method ESR-O2O with vanilla SR-based methods and show the differences in the framework in Figure \ref{fig:frame}. Vanilla SR-base methods often assume access to multiple source environments to learn representations. In such cases, the generalization to target environments can be considered as the interpolation \cite{SRnavigation2017} or modeling the Gaussian process \cite{sfde2021}. However, our method faces the more challenging task of task generalization with access to only an offline dataset generated from a single environment. On the other hand, vanilla methods typically assume a linear relationship between the value function and the successor representation, which restricts the applicability of successor representations. In contrast, our method directly models the value function as a function of successor representations, utilizing multiple layers rather than a single linear layer. This broader modeling capability enhances the feasibility and applications of our method in more complex scenarios.

Formally, we introduce the ensemble successor representations and ensemble Q functions as follows. As depicted in Figure \ref{fig:frame}(b), $\psi_i: i\in[1,n], |\mathcal{S}|\times|\mathcal{A}|\rightarrow |\mathcal{S}|$ represents the ensemble successor representations, where each member is initialized randomly. Similarly, $Q_k: k\in[1,n], |\mathcal{S}|\times|\mathcal{A}|\rightarrow 1$ represents the ensemble Q functions. Both the ensemble SR and ensemble Q functions are updated using TD learning:
\begin{equation}
\psi_k(s_t,a_t) \leftarrow \psi_k(s_t,a_t) + \alpha [\phi(s_t,a_t) + \gamma\cdot \psi_k(s_{t+1},a_{t+1}) - \psi_k(s_t,a_t)]\label{eq:Phi_k},
\end{equation}
\begin{equation}
Q_k(\psi_k(s_t,a_t)) \leftarrow Q_k(\psi_k(s_t,a_t))+\alpha[r(s_t,a_t) +\gamma\cdot Q_k(\psi_k(s_{t+1},a_{t+1}))-Q_k(\psi_k(s_t,a_t))],\label{eq:Q_k}
\end{equation}
where $\alpha$ and $\gamma$ represent the learning rate and discount factor, respectively. The basic feature $\psi(\cdot)$ is assumed to be available in this work. When the input is finite and vectored, this mapping can be simplified as the identity function. The input of the Q functions is the successor representation, which differs from linear composition (i.e., $Q=\psi^\top w$). It is noted that we utilize independent targets instead of shared targets, which may introduce optimism in certain cases \cite{msg2022}.

\begin{algorithm}[ht]
\footnotesize
\caption{Learning representation, policy, and value function from offline data}
\label{alg:offline}
\begin{algorithmic}[1]
\REQUIRE Offline dataset $D$ generated from $M_i$, policy network $\pi$, $Q$-networks $\{Q_{k}\}_{k\in[n]}$, and target $Q$-networks $\{\bar{Q}_{k}\}_{k\in[n]}$, representation networks $\{\psi_{k}\}_{k\in[n]}$, and target representation networks $\{\bar{\psi}_{k}\}_{k\in[n]}$.
\STATE Initialize the parameters of $\pi$, $\{Q_{k}\}_{k\in[n]}$, $\{\bar{Q}_{k}\}_{k\in[n]}$, $\{\psi_{k}\}_{k\in[n]}$, and $\{\bar{\psi}_{k}\}_{k\in[n]}$.
\WHILE{\emph{not coverage}}
\STATE Sample transitions $\{(s,a,r,s')\}$ from $D$
\STATE Update successor representations with Eq. (\ref{eq:Phi_k})
\STATE Update critic networks using Eq. (\ref{eq:Q_k})
\STATE Update target networks using Eq. (\ref{eq:target_net})
\STATE Update the policy network greedily according to Eq. (\ref{eq:policy})
\ENDWHILE
\end{algorithmic}
\end{algorithm}
\subsection{Offline pre-training and Fine-tuning Process}

In this section, we present the offline pre-training process and online fine-tuning process. Algorithm \ref{alg:offline} outlines our approach. During the offline training stage, the representations and the critic network are trained based on the Bellman equation. In each mini-batch, all networks in the ensemble are updated using Equations (\ref{eq:Phi_k}) and (\ref{eq:Q_k}). Target networks are used to stabilize the learning process. If we define the respective parameters of the original network and the target network as $\theta$ and $\bar{\theta}$, then target networks can be updated using Polyak averaging:
\begin{equation}
    \bar{\theta} \leftarrow \rho \bar{\theta} + (1-\rho) \theta.
    \label{eq:target_net}
\end{equation}
The policy network is trained to maximize the minimum value among the ensemble of critics, which can be formulated as follows:
\begin{equation}
    \pi(\cdot|s) \leftarrow \arg\max_a \min_k Q_k(s,a) = \arg\max_a \min_k Q_k(\psi_k(s,a)).
    \label{eq:policy}
\end{equation}
This can be thought of as forming a lower confidence bound (LCB) for the value function of a policy using the batch data and then seeking to find a policy that maximizes the LCB. 

\begin{algorithm}[t]
\footnotesize
\caption{Online fine-tuning process given pre-trained ESR}
\label{alg:online}
\begin{algorithmic}[1]
\REQUIRE Online environment $M_j$, fine-tuning steps $T$;
\STATE Load pre-trained policy network $\pi$, pre-trained critic networks $Q^{\pi}_i$, target networks $\bar{Q}^{\pi}_i$, and learned SR $\psi^{\pi}$;
\STATE Fix the parameters of representation networks $\psi^{\pi}$;
\REPEAT
\STATE Interact with the environment $M_j$ with action $a\sim \pi(\cdot|s)$;
\STATE Receive new state $s'$ and reward $r$ from the environment;
\STATE Store transitions $\{s,a,r,s'\}$ into replay buffer;
\STATE Update critic networks and policies according to Eq. (\ref{eq:Q_k}) and (\ref{eq:policy});
\STATE Update target networks
\UNTIL{online steps $=T$}
\end{algorithmic}
\end{algorithm}

When it comes to online fine-tuning, ESR-O2O loads all pre-trained networks, including the policy network, pre-trained critic networks, target networks, and the representation network. We do not set many fine-tuning steps, since large number of online interactions are not available. To prevent the representation from being compromised or even destroyed \cite{pex2023}, we fix the parameters of representation networks during the fine-tuning process. Algorithm \ref{alg:online} outlines our online fine-tuning procedure.

\section{Experiments}

In this section, we present experimental evaluations to assess the effectiveness and feasibility of our proposed method. Specifically, we aim to address the following research questions: 1) Does the utilization of ensembles for successor representations and Q functions lead to improved performance in both the offline learning and fine-tuning stages? 2) In scenarios without reward gaps, does ensemble SR outperform existing offline-to-online learning approaches? 3) What is the significance of ensembles in achieving superior performance with our method? 

\begin{figure}[ht]
    \centering
    \subfloat[Reach]
    {\includegraphics[width=0.12\linewidth,height=0.12\linewidth]{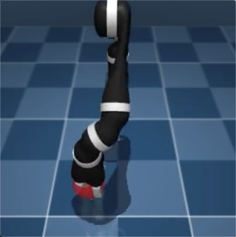}
    \includegraphics[width=0.12\linewidth,height=0.12\linewidth]{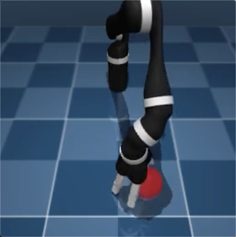}}
    \quad
    \subfloat[Quadruped]{\includegraphics[width=0.12\linewidth,height=0.12\linewidth]{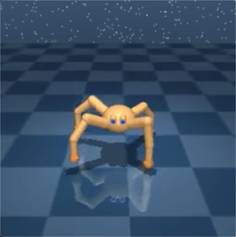}
    \includegraphics[width=0.12\linewidth,height=0.12\linewidth]{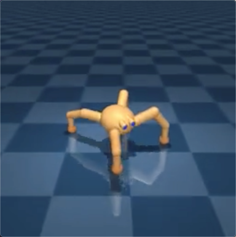}}
    \quad
    \subfloat[Walker]{\includegraphics[width=0.12\linewidth,height=0.12\linewidth]{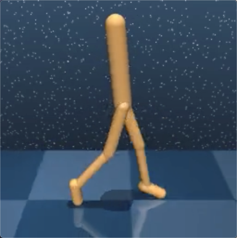}
    \includegraphics[width=0.12\linewidth,height=0.12\linewidth]{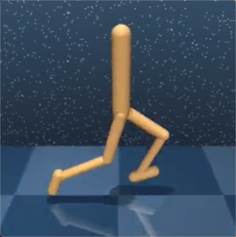}
    \includegraphics[width=0.12\linewidth,height=0.12\linewidth]{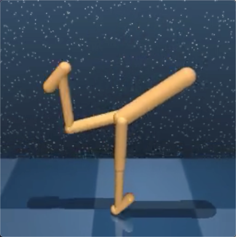}}

    \vspace{-1em}
    \subfloat[MetaWorld]
    {\includegraphics[width=0.12\linewidth,height=0.12\linewidth]{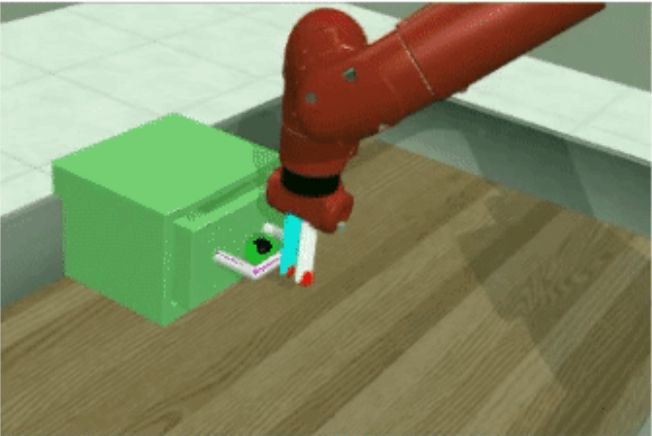}
    \includegraphics[width=0.12\linewidth,height=0.12\linewidth]{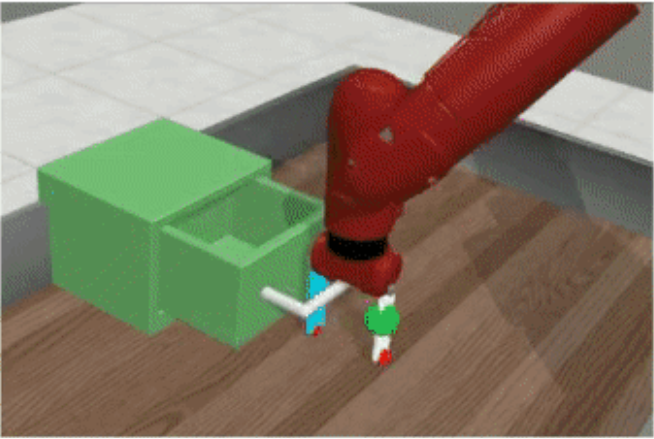}
    \quad
    \includegraphics[width=0.12\linewidth,height=0.12\linewidth]{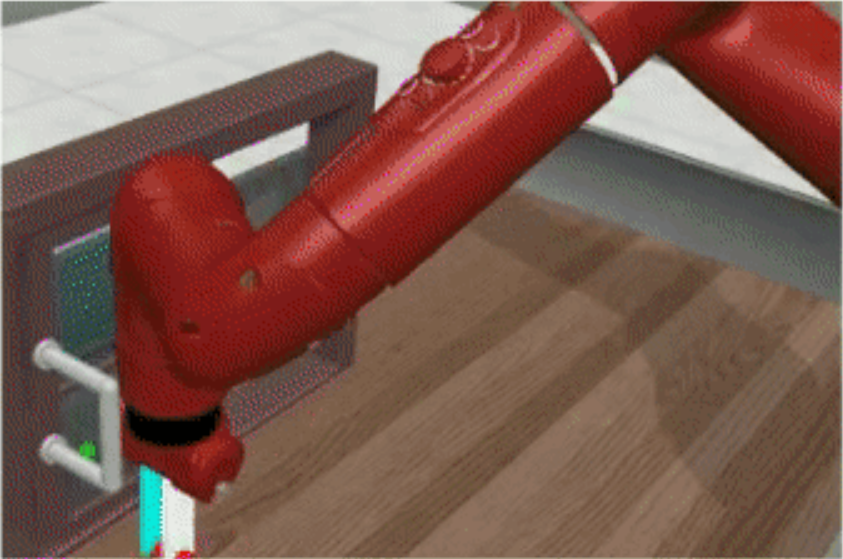}
    \includegraphics[width=0.12\linewidth,height=0.12\linewidth]{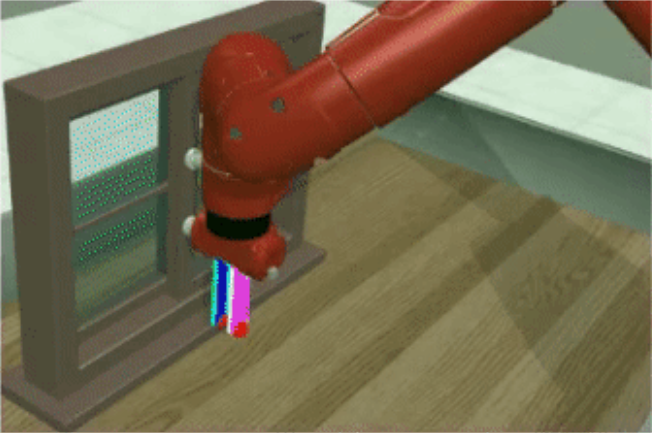}}
    \quad
    \subfloat[AntMaze]{\includegraphics[width=0.12\linewidth,height=0.12\linewidth]{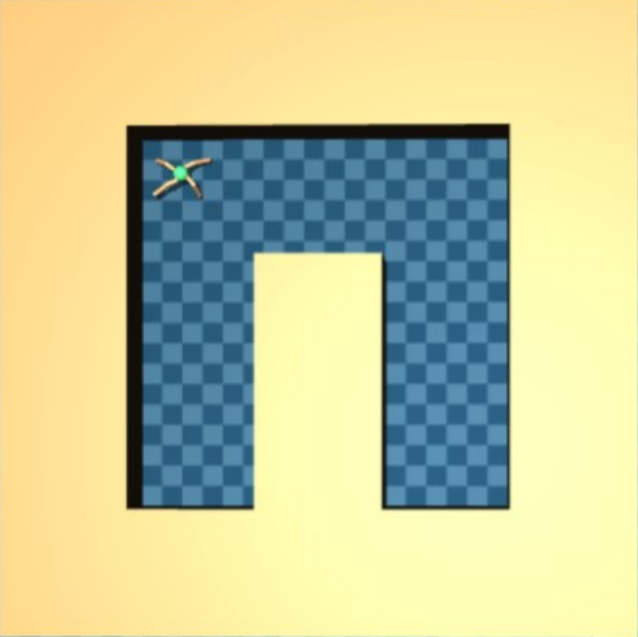}
    \includegraphics[width=0.12\linewidth,height=0.12\linewidth]{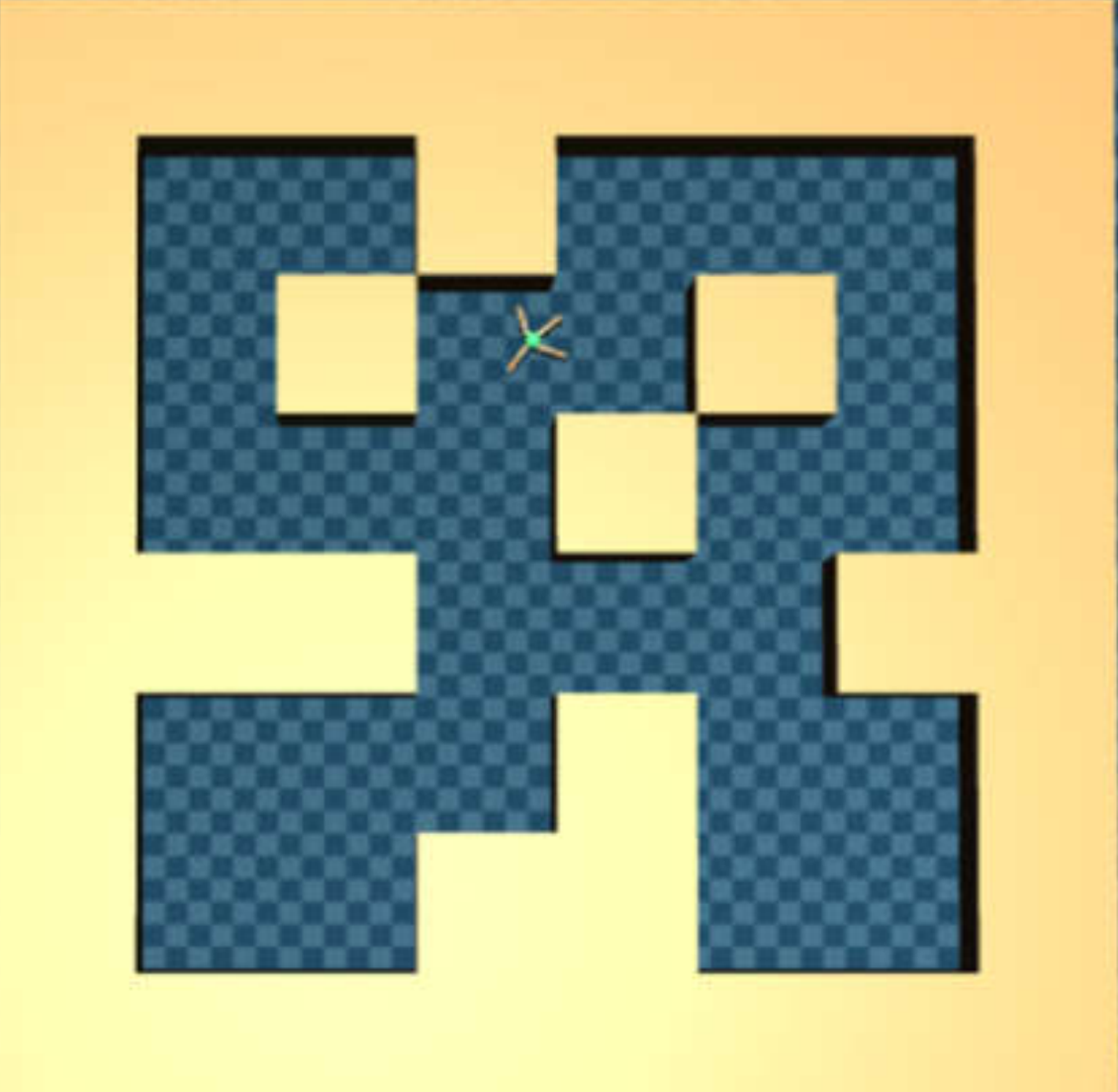}
    \includegraphics[width=0.12\linewidth,height=0.12\linewidth]{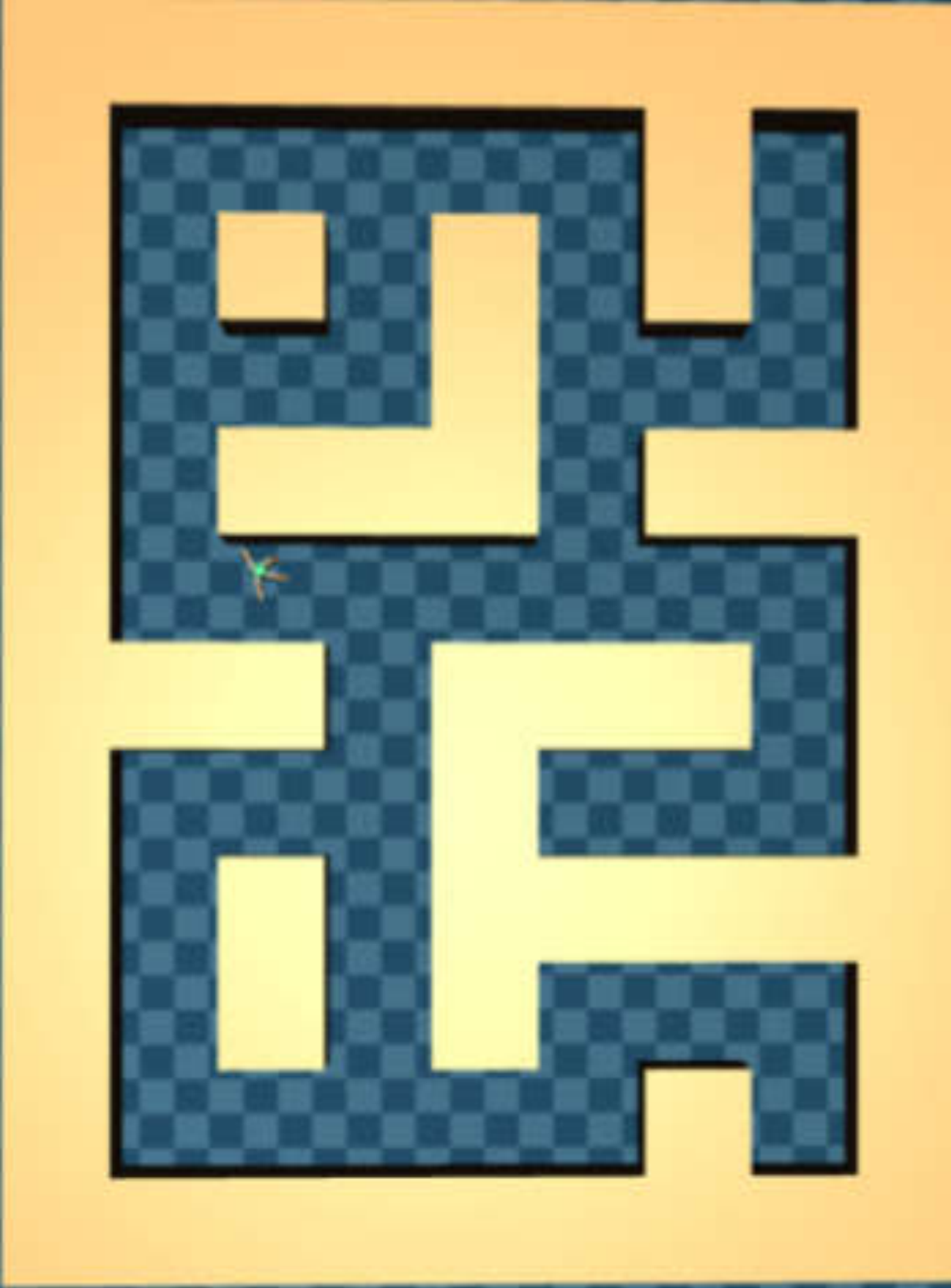}}
    \caption{Experimental environments, including Quadruped, Walker, Reach from UTDS \cite{utds2023}, MetaWorld \cite{meta2020}, and Antmaze from D4RL \cite{d4rl2020}. These environments can be categorized into three classes based on the magnitude of the reward gap: small gap (Quadruped, Walker), big gap (Reach, MetaWorld), and no gap (Antmaze).}
    \label{fig:envs}
\end{figure}
\subsection{Setups}

All experimental environments are illustrated in Figure \ref{fig:envs}. The Quadruped environment involves tasks such as walking, running, jumping, and rolling, while the Walker environment focuses on walking, running, and flipping tasks. In the Reach environment, a manipulator is required to place blocks in different positions, including bottom left, bottom right, top left, and top right. Each goal corresponds to a distinct reward function. To investigate task generalization during fine-tuning from a single offline environment, we randomly select two tasks, denoted as $M_i$ and $M_j$, from the aforementioned environments. Agents pre-trained on the offline dataset generated in $M_i$ are then fine-tuned and evaluated in the new environment $M_j$, which shares the same dynamics but possesses a different reward function. We refer to Appendix A for more details.

\begin{figure}[!t]
\centering
\includegraphics[width=\linewidth]{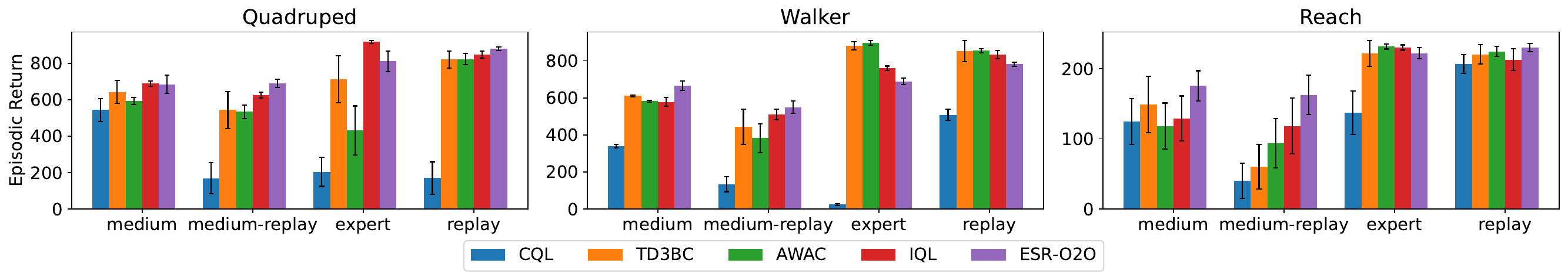}
\caption{Offline Performance after 1M training steps. Each bar represents the performance averaged on different tasks in the same environments. The error bar are estimated across 5 random seeds.}
\label{fig:offline_performance}
\end{figure}

\subsection{Performance when reward functions change}

Due to the absence of available baselines specific to our setting, we incorporate several offline-to-online learning methods like AWAC \cite{AWAC2020}, Off2On \cite{off2onrl2022}, PEX \cite{pex2023}, and PROTO \cite{proto2023} as baselines for comparison. We also compare our method with offline RL methods such as CQL \cite{cql2020}, IQL \cite{iql2022}, TD3BC \cite{td3bc2021}, and present their fine-tuning results in Appendix B due to page limitations. To assess the performance, we categorize the environments based on the magnitude of the reward gap between $M_i$ and $M_j$, as shown in Figure \ref{fig:envs}. When the reward gap is small, previously learned policies and value functions may be effectively utilized, implying easier generalization. In contrast, harder generalization scenarios involve a large reward gap, making it challenging for the learned policies to perform well in the new task.

\paragraph{Offline performance.} We begin by comparing the performance of pre-trained agents using various algorithms. As illustrated in Figure \ref{fig:offline_performance}, our method outperforms the start-of-the-art offline RL algorithms with `medium' and `medium-replay' data, and exhibit competitive performance given `expert' and `replay' data. Furthermore, our method achieves reduced variances in episodic returns compared to baseline algorithms, showing enhanced stability. We also observe that the performance gaps across different datasets, such as the disparity between `replay' data and `medium-replay' data, are more pronounced in baseline methods, whereas our proposed method displays narrower performance gaps among these datasets. This implies that ESR-O2O is more robust to the data coverage and quality of offline datasets.

\begin{figure}[!t]
\centering
\includegraphics[width=0.9\linewidth]{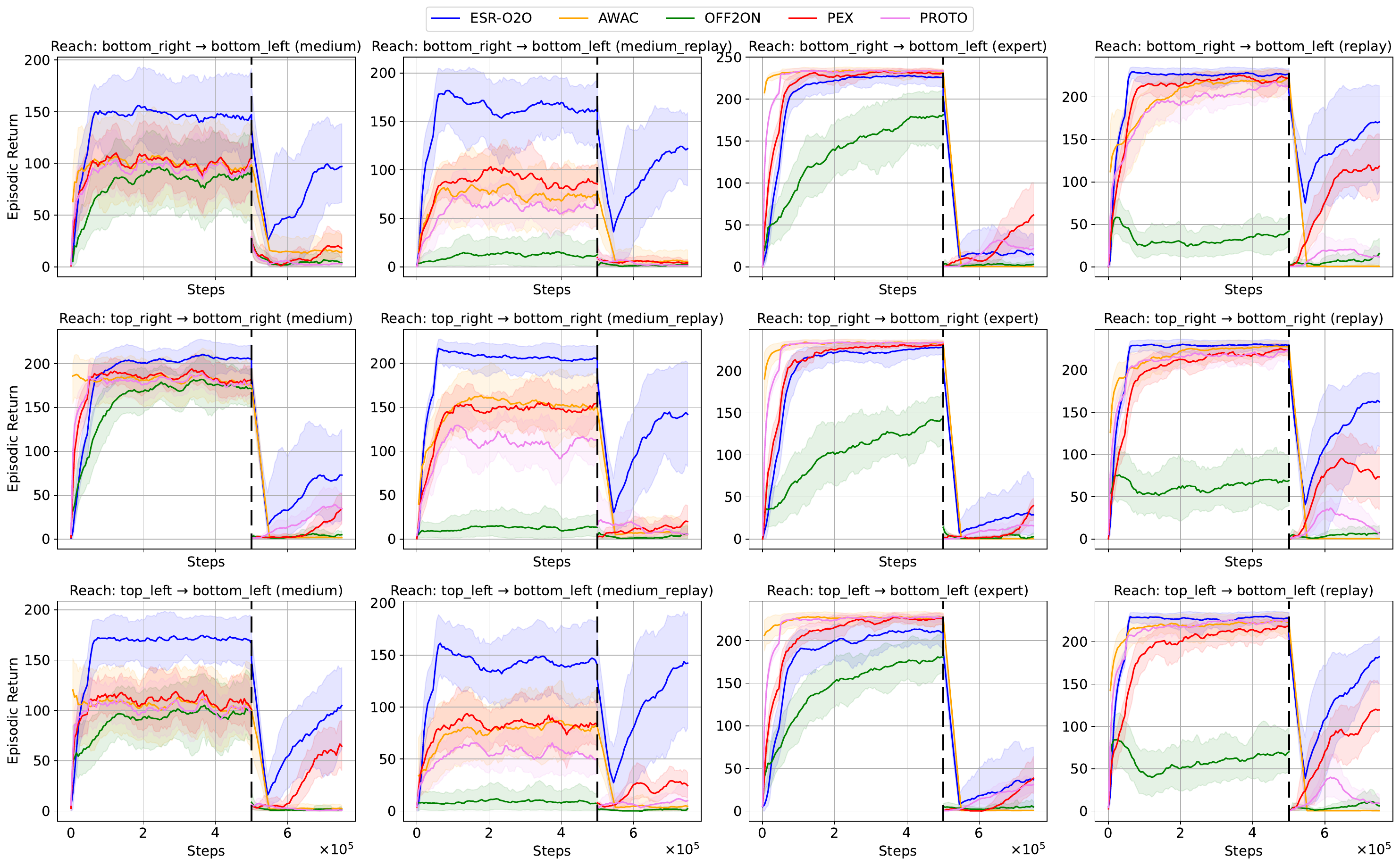}
\caption{Fine-tuning performance in scenarios with large reward gaps. While baseline methods struggle to perform well during the fine-tuning stage, ESR-O2O demonstrates significant improvements in the fine-tuned policies, regardless of the quality of the offline datasets.}
\label{fig:biggap}
\end{figure}

\paragraph{Fine-tuning performance with a big reward gap.} In scenarios with a significant reward gap, previously learned policies or value functions become ineffective. As presented in Figure \ref{fig:biggap}, conventional methods fail to acquire useful policies in the fine-tuning stage. However, ESR-O2O demonstrates the ability to effectively handle such reward gaps and achieve robust task generalization. The performance of ESR-O2O on `medium-replay' data closely approaches that of `replay' data, indicating its capacity to learn effectively even from datasets lacking expert policies. When `expert' data is provided, ESR-O2O exhibits favorable performance during offline training, albeit the fine-tuning performance is inferior to those on other types of offline data. This suggests that in the presence of a highly narrow dataset, the ability of ESR-O2O to diversify representations and Q functions and improve performance through limited interactions may be constrained. PEX also demonstrate adaptability in several tasks due to its adaptive policy composition, but still falls short of our method's performance. We also present the comparisons in the MetaWorld benchmark in Table \ref{tab:meta}. These results also validate the superiority of our method when fine-tuning in a new environment, with more detailed results available in Appendix B.

\begin{table}[ht]
\caption{Average return on MetaWorld tasks, the best results in online fine-tuning are bold.}
\label{tab:meta}
\resizebox{\textwidth}{!}{
\begin{tabular}{lccc}
\toprule
Environments & PEX & PROTO & ESR-O2O\\ 
\midrule
drawer-close-v2 $\rightarrow$ drawer-open-v2  & 3409.04 $\pm$ 1702.83 $\rightarrow$ 812.49 $\pm$ 789.24 & 3646.20 $\pm$ 885.18 $\rightarrow$ 395.71 $\pm$ 29.85  & 3249.58 $\pm$ 1874.60 $\rightarrow$ \textbf{2278.42} $\pm$ 275.47\\
window-open-v2 $\rightarrow$ window-close-v2  & 394.35 $\pm$255.25 $\rightarrow$ 571.73 $\pm$369.45 & 977.18 $\pm$ 870.05 $\rightarrow$ 174.69 $\pm$ 164.64 & 280.81 $\pm$ 15.8604 $\rightarrow$ \textbf{2910.27} $\pm$ 1446.12\\
\bottomrule
\end{tabular}}
\end{table}

\vspace{-1em}

\begin{figure}[!t]
\centering
\includegraphics[width=0.9\linewidth]{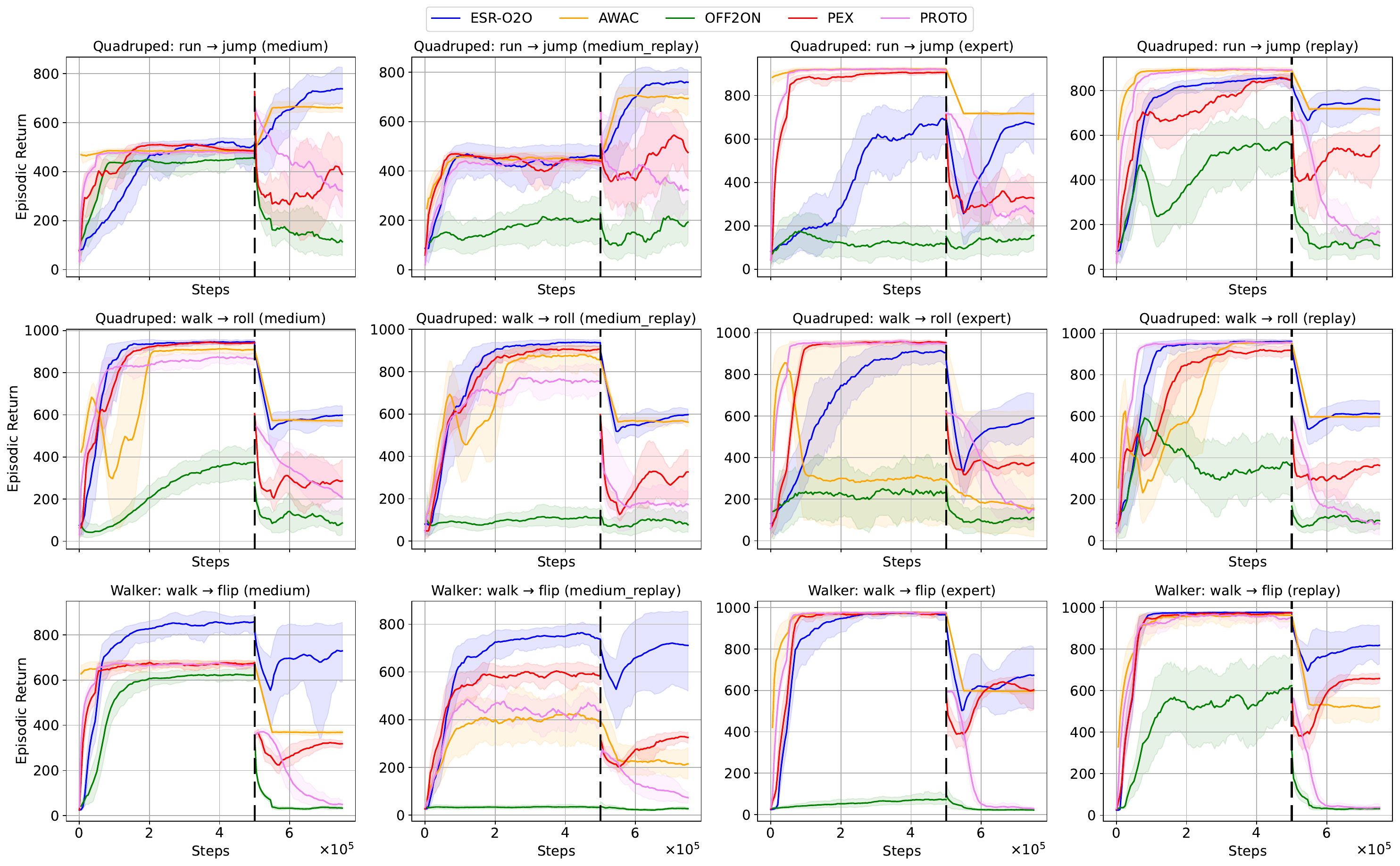}
\caption{Fine-tuning performance comparisons in scenarios with small reward gaps. The shaded areas indicate variances across 5 random seeds.}
\label{fig:smallgap}
\end{figure}

\paragraph{Fine-tuning Performance with a small reward gap.} In scenarios with small reward gaps, it is possible to leverage pre-trained policies or value functions for the new task. PEX and AWAC demonstrate a degree of robustness in task switching. Conversely, Off2On, which is built on top of CQL and extracts policies by maximizing Q-functions, appears to be more sensitive to the distributional shift, resulting in instability. Figure \ref{fig:smallgap} illustrates that ESR-O2O significantly outperforms other methods. While PEX and PROTO swiftly acquire useful policies during pre-training, we observe performance degradation and inferiority during fine-tuning in a new environment. We speculate this is due to the lack of consideration for reward gaps and unsuitable data sampling methods.

\begin{table}[ht]
\caption{Performance on Antmaze tasks, the best scores in offline pre-training and online fine-tuning are bold.}
\label{tab:antmaze}
\resizebox{\textwidth}{!}{
\begin{tabular}{lccccc}
\toprule
Environments & AWAC & CQL & IQL &  Cal-QL & ESR-O2O\\ 
\midrule
antmaze-umaze-v2 & 52.75 $\pm$8.67 $\rightarrow$ 98.75 $\pm$1.09  &  94.00 $\pm$1.58 $\rightarrow$ 99.50 $\pm$0.87  & 77.00 $\pm$0.71 $\rightarrow$ 96.50 $\pm$1.12 & 76.75 $\pm$ 7.53 $\rightarrow$ \textbf{99.75} $\pm$ \textbf{0.43}  & \textbf{98.00} $\pm$ \textbf{1.87} $\rightarrow$ 99.2 $\pm$ 1.79\\
antmaze-umaze-diverse-v2  & 56.00 $\pm$2.74 $\rightarrow$ 0.00 $\pm$0.00  & 9.50 $\pm$9.91 $\rightarrow$ \textbf{99.00} $\pm$ \textbf{1.22} & 59.50 $\pm$9.55 $\rightarrow$ 63.75 $\pm$25.02 & 32.00 $\pm$ 27.79 $\rightarrow$ 98.50 $\pm$ 1.12 & \textbf{93.75} $\pm$ \textbf{2.63} $\rightarrow$ 98.75 $\pm$ 0.96\\
antmaze-medium-play-v2  & 0.00 $\pm$0.00 $\rightarrow$ 0.00 $\pm$0.00 & 59.00 $\pm$11.18 $\rightarrow$ 97.75 $\pm$1.30 & 71.75 $\pm$2.95 $\rightarrow$ 89.75 $\pm$1.09 & 71.75 $\pm$ 3.27 $\rightarrow$ \textbf{98.75} $\pm$ \textbf{1.64} & \textbf{76.00} $\pm$ \textbf{1.41} $\rightarrow$ 97.00 $\pm$ 2.83\\
antmaze-medium-diverse-v2 & 0.00 $\pm$0.00 $\rightarrow$ 0.00 $\pm$0.00 & 63.50 $\pm$6.84 $\rightarrow$ 97.25 $\pm$1.92 & \textbf{64.25} $\pm$ \textbf{1.92} $\rightarrow$ 92.25 $\pm$2.86 & 62.00 $\pm$ 4.30 $\rightarrow$ 98.25 $\pm$ 1.48 & 59.40 $\pm$ 17.83 $\rightarrow$ \textbf{98.6} $\pm$ \textbf{1.67}\\
antmaze-large-play-v2 & 0.00 $\pm$0.00 $\rightarrow$ 0.00 $\pm$0.00 & 28.75 $\pm$7.76 $\rightarrow$ 88.25 $\pm$2.28 & 38.50 $\pm$8.73 $\rightarrow$ 64.50 $\pm$17.04 & 31.75 $\pm$ 8.87 $\rightarrow$ 97.25 $\pm$ 1.79 & \textbf{71.6} $\pm$ \textbf{5.32} $\rightarrow$ \textbf{97.8} $\pm$ \textbf{1.30}\\
antmaze-large-diverse-v2 & 0.00 $\pm$0.00 $\rightarrow$ 0.00 $\pm$0.00 & 35.50 $\pm$3.64 $\rightarrow$ 91.75 $\pm$3.96 & 26.75 $\pm$3.77 $\rightarrow$ 64.25 $\pm$4.15 & 44.00 $\pm$ 8.69 $\rightarrow$ 91.50 $\pm$ 1.79 & \textbf{73.4} $\pm$ \textbf{6.35} $\rightarrow$ \textbf{97.8} $\pm$ \textbf{1.48}\\
\midrule
Average & 18.12 $\rightarrow$ 16.46  & 48.38 $\rightarrow$ 95.58 & 56.29 $\rightarrow$ 78.50 & 53.04 $\rightarrow$ 97.33 & \textbf{78.69} $\rightarrow$ \textbf{98.19}\\ 
\bottomrule
\end{tabular}}
\end{table}

\subsection{Performance when rewards do not change}
We also conduct experiments to compare ESR-O2O with other offline-to-online methods that focus on the fine-tuning process within a single task. For this purpose, we choose the challenging navigation task Antmaze from D4RL, which is a widely-used benchmark that do not contain reward changes. The difficulty of this tasks lies in the sparse rewards and the need of exploration. Baselines including CQL, IQL, AWAC, and Cal-QL \cite{calql2023} are adopted. The results of the comparisons are presented in Table \ref{tab:antmaze}. Among the baselines, IQL shows the best offline performance, while Cal-QL achieves the best fine-tuning performance on most tasks. However, ESR-O2O outperforms all these methods by a significant margin, especially in the case of `antmaze-large' tasks with the most complex environments.

\begin{figure}[htbp]
\centering
\includegraphics[width=0.9\linewidth]{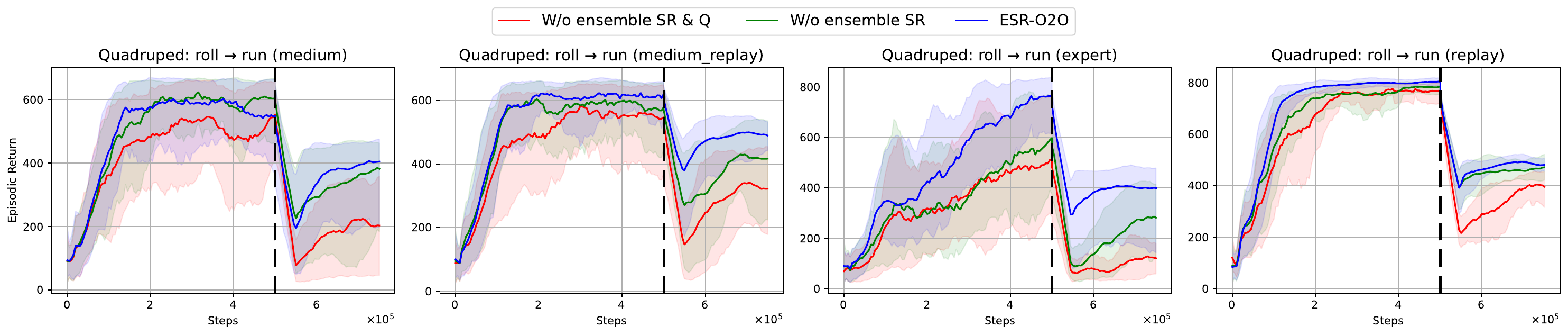}
\includegraphics[width=0.9\linewidth]{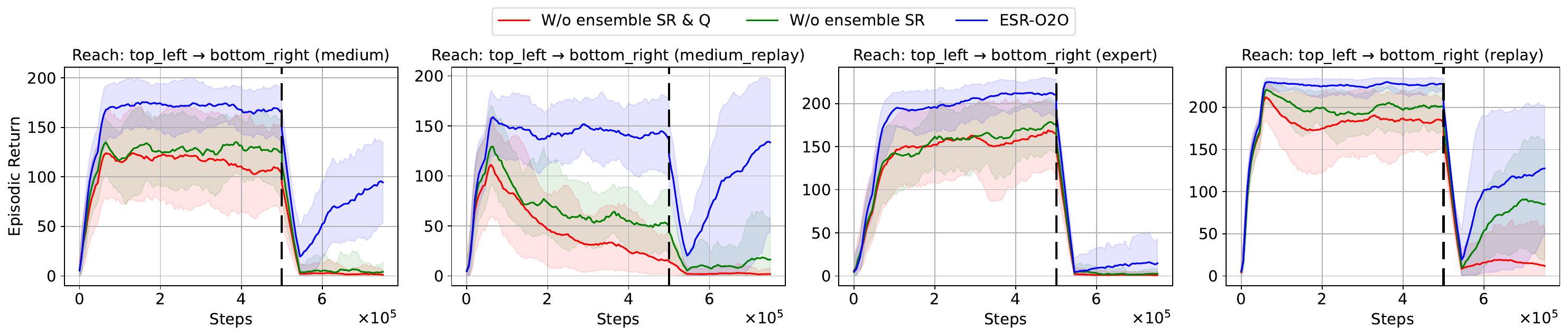}
\caption{Ablation on ensemble SR and ensemble Q. The shaded area indicates the variance across 5 random seeds.}
\label{fig:ablation}
\end{figure}
\vspace{-2em}
\begin{figure}[ht]
\centering
\includegraphics[width=0.9\linewidth]{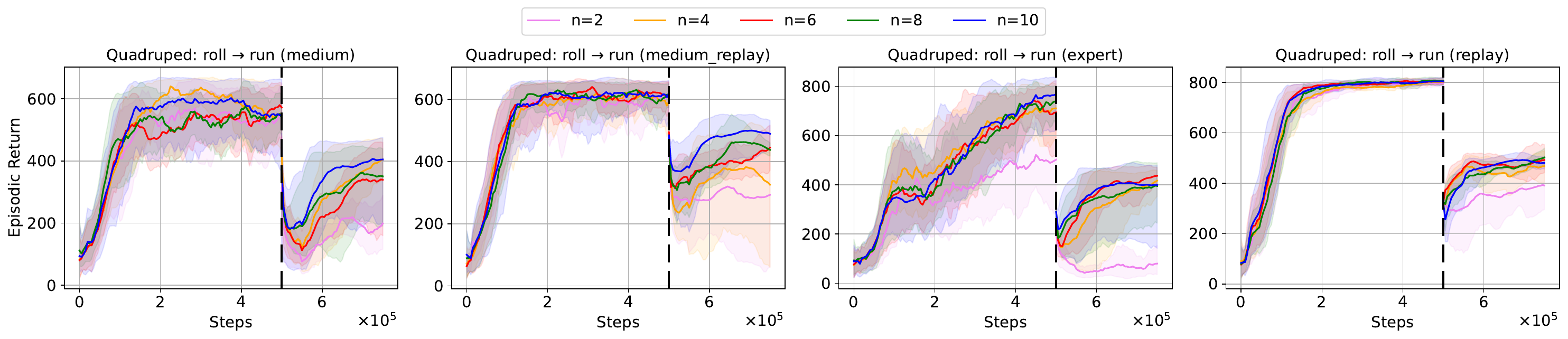}
\includegraphics[width=0.9\linewidth]{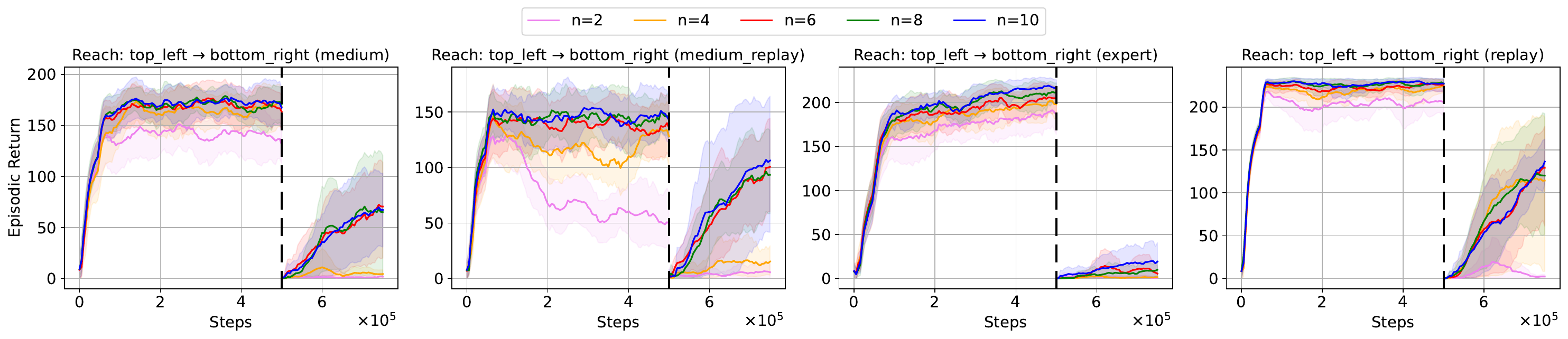}
\caption{Ablation on the quantity of ensemble networks $n$. The shaded area indicates the variance across 3 random seeds.}
\label{fig:ablation2}
\end{figure}

\vspace{-1em}
\subsection{Ablation study}
To demonstrate the feasibility of our proposed design, we examine the impact of ensembles applied on SR and Q networks on the performance. Firstly, we compare our method with two variants by eliminating ensembles of SR (indicated by `W/o ensemble SR') and all ensembles (indicated by `W/o ensemble SR $\&$ Q'). In the variant without ensemble SR and Q, ESR-O2O essentially becomes the original version of successor representation as describe in Section \ref{sec:motivating}. This variant only incorporates Layer Normalization to mitigate overestimation during offline pre-training. Secondly, we evaluate the effect of the number of ensemble networks when $n$ range from 2 to 10. We present the ablation results in Figures \ref{fig:ablation} and \ref{fig:ablation2}. These results illustrate that the ensemble of SR is critical for fine-tuning, especially for large reward gaps. The ensemble of Q networks also plays a significant role when dealing with scenarios with a small reward gap or when the offline data is abundant. As for the quantity of ensemble networks, we observe that our method exhibit strong performance when $n$ is 6 or greater. We refer to Appendix C for more details.

\vspace{-1em}
\section{Conclusion}

In this paper, we have proposed a novel method incorporating ensemble networks and successor representations to handle the task generalization problem in the offline-to-online RL setting. The integration of ensemble networks allows our model to capture diverse representations and reward functions of the environment. Through extensive experiments on various benchmark tasks, we have demonstrated significant improvements in the agent's ability to transfer knowledge from offline data to online environments. Our method outperforms state-of-the-art techniques and provides superior generalization performance. Our work contributes to the broader field of RL by addressing a fundamental limitation of offline-to-online learning. The proposed method opens up new avenues for real-world applications where collecting online data is expensive or time-consuming, with the potential to handle task generalization.

In the fine-tuning phase, we have chosen to fix the representation networks to prevent deviation caused by distributional shifts. However, it is worth exploring the possibility of fine-tuning the representation networks using new experiences. Techniques like model expansion \cite{pex2023} could also be employed to further stabilize the representations in the face of distributional shifts. Additionally, investigating more effective measurement for reward gaps is important. KL or JS divergences can be used to characterize the gap between reward distributions, and diffusion models can serve as an efficient distribution estimator \cite{diffusion_distribution}.

\vspace{-1em}

\section*{Acknowledgments}
This work is supported by the National Science Fund for Distinguished Young Scholars (Grant No.62025602), the National Natural Science Foundation of China (Grant Nos. 62306242, U22B2036, 11931015), Fok Ying-Tong Education Foundation China (No.171105), the Tencent Foundation, XPLORER PRIZE, the Science Center Program of National Natural Science Foundation of China under Grant 62188101, and the Heilongjiang Touyan Innovation Team Program.

\clearpage
\begin{appendix}
\section{Implementation Details.}

\paragraph{Environments.} In our experiments, except for Mujoco style tasks, we also incorporate MetaWorld tasks and AntMaze navigation tasks. The MetaWorld benchmark is paramount in multi-task reinforcement learning. Antmaze tasks from D4RL are widely used in offline RL and offline-to-online RL settings. Although the Antmaze environment does not involve reward changes, it presents significant challenges in terms of exploration and sparse rewards. We list the choice of $M_i$ and $M_j$ in the experimental section below.

\begin{table}[htbp]
\footnotesize
\centering
\caption{All combinations of $M_i$ and $M_j$ in Section 6.}
\label{tab:tasks}
\begin{tabular}{ccc}
\hline
\textbf{Environments}      & \textbf{$M_i$ (offline task)} & \textbf{$M_j$ (online task)} \\ \hline
Quadruped                  & walk                         & roll                        \\ \hline
\multirow{2}{*}{Walker}    & run                          & jump                        \\
                           & walk                         & flip                        \\ \hline
\multirow{3}{*}{Reach}     & top left                     & bottom left                 \\
                           & top right                    & bottom right                \\
                           & bottom right                 & bottom left                 \\ \hline
\multirow{2}{*}{MetaWorld} & drawer-close-v2              & drawer-open-v2              \\
                           & window-open-v2               & window-close-v2             \\ \hline
\multirow{3}{*}{Antmaze}   & umaze                        & umaze                       \\
                           & medium                       & medium                      \\
                           & large                        & large                       \\ \hline
\end{tabular}
\end{table}

\paragraph{Datasets.} The offline datasets in our experiments for the Quadruped, Walker, and Reach environments are obtained from UTDS \cite{utds2023}. UTDS employs the TD3 \cite{td32018} algorithm to collect data at different levels. The dataset consists of four types: medium-replay, medium, expert, and replay. The medium-replay data encompasses all experiences collected during the training of a medium-level TD3 agent, while the replay data includes all experiences obtained from training an expert-level TD3 agent. The medium and expert data denote narrower data distributions, where the medium data is generated by a medium-level TD3 agent and the expert data is generated by an expert-level TD3 agent. The AntMaze dataset is obtained from D4RL \cite{d4rl2020}, a widely used resource in offline RL. The offline dataset of MetaWorld tasks is obtained through interactions with the officially provided scripted policy.

\paragraph{Implementation of our method and offline RL methods.} For Quadruped, Walker, and Reach tasks, we use the Jax implementations of baseline methods from jrlzoo\footnote{https://github.com/fuyw/jrlzoo}. We build the implementation of ESR-O2O on top of rlpd\footnote{https://github.com/ikostrikov/rlpd}. For Antmaze tasks, we cite the reported scores from CORL\footnote{https://github.com/tinkoff-ai/CORL}. Agents are pre-trained for 1M steps and then fine-tuned for 250k steps. Each method on each task has been run across 5 random seeds. In Antmaze environments, since behavior cloning is commonly employed in other offline-to-online learning methods, we incorporate it into actor learning as well.

\paragraph{Implementations of offline-to-online RL methods.} In our work, we compare ESR-O2O with many offline-to-online RL algorithms, including AWAC, Off2On, PEX, and PROTO, though they are designed for online fine-tuning without reward gaps. For Off2On, PEX, and PROTO, we build on top of their official codes, with light modifications to be suitable for our tasks. We use their default architectures and hyper-parameters. Agents are pre-trained for 500k timesteps during the offline stage and then fine-tuned for 250k timesteps in the new environment.

\clearpage
\section{}
\subsection{Comparisons with offline RL methods.}
\begin{figure}[ht]
\centering
\includegraphics[width=0.7\linewidth]{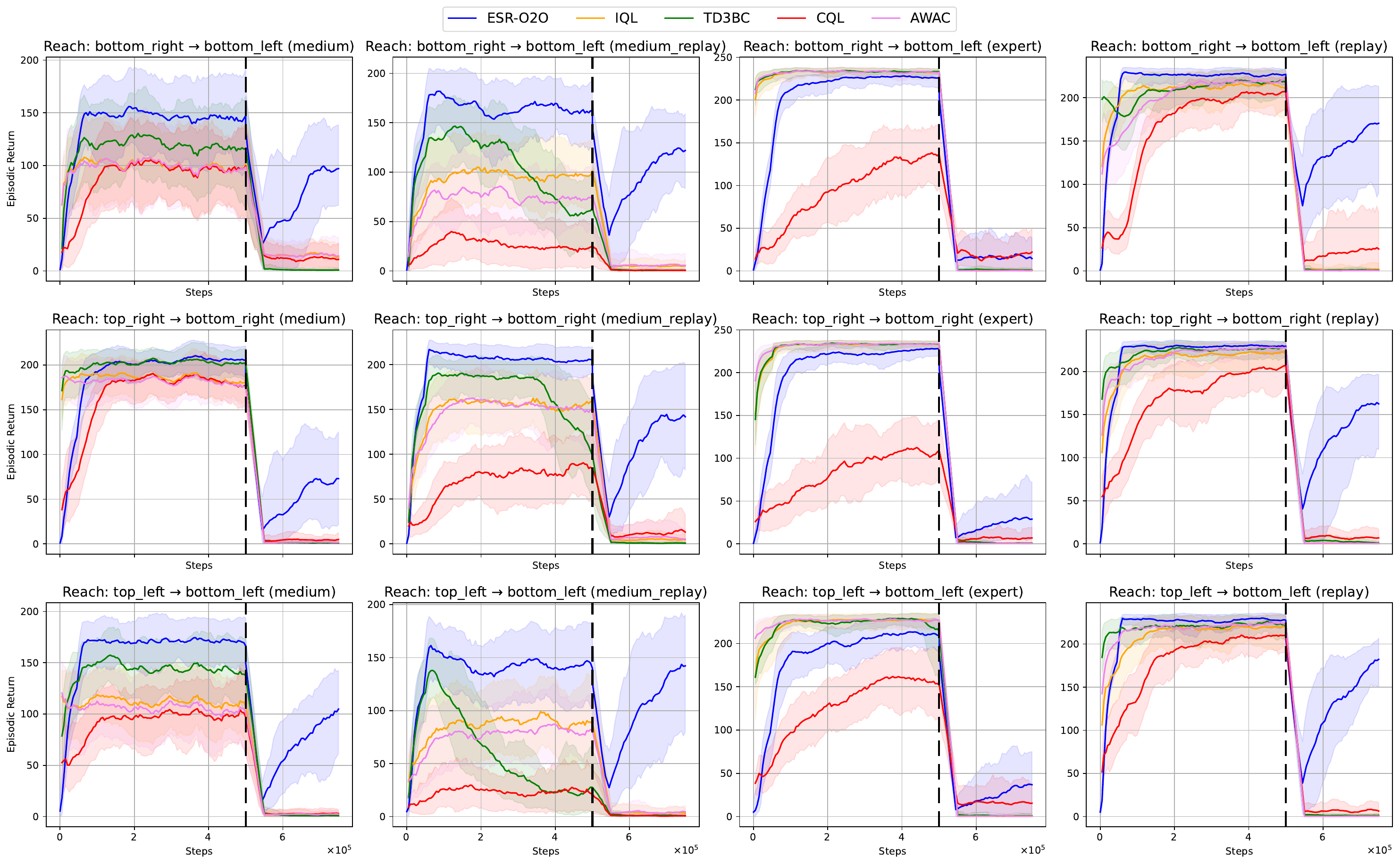}
\caption{Fine-tuning performance in scenarios with large reward gaps. While baseline methods struggle to perform well across all tasks during the fine-tuning stage, ESR-O2O demonstrates significant improvements in the fine-tuned policies, regardless of the quality of the offline datasets.}
\label{fig:biggap_offline}
\end{figure}

\begin{figure}[ht]
\centering
\includegraphics[width=0.7\linewidth]{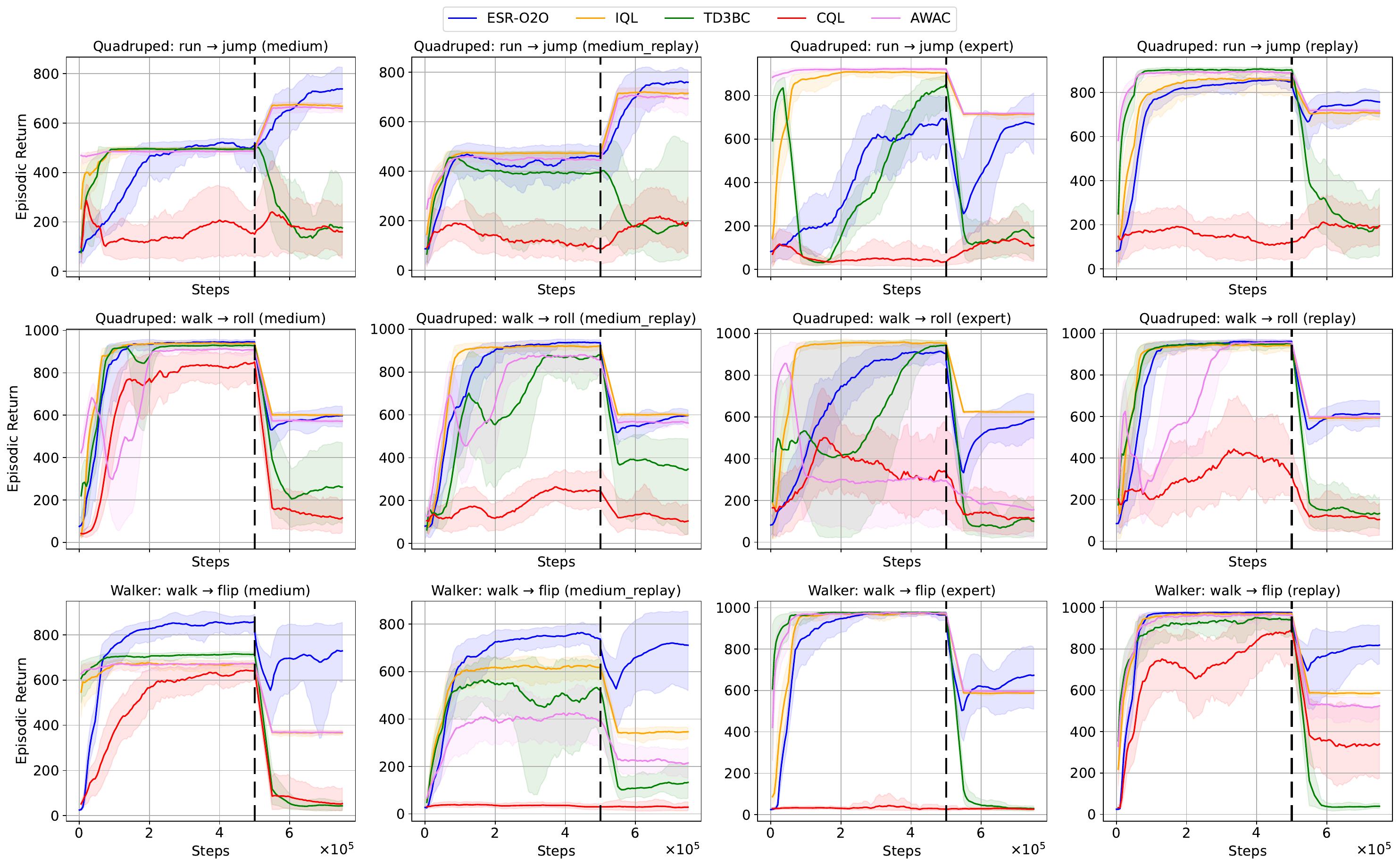}
\caption{Fine-tuning performance in scenarios with large reward gaps. While baseline methods struggle to perform well across all tasks during the fine-tuning stage, ESR-O2O demonstrates significant improvements in the fine-tuned policies, regardless of the quality of the offline datasets.}
\label{fig:smallgap_offline}
\end{figure}

When the reward gap is large, all offline RL methods cannot obtain effective policies during fine-tuning, as shown in Figure \ref{fig:biggap_offline}. In scenarios with small reward gaps, it is possible to leverage pre-trained policies or value functions for the new task. IQL and AWAC, which employ the same policy extraction method, demonstrate robustness in task switching. Conversely, CQL and TD3-BC, which extract policies by maximizing the Q-functions, are more sensitive to the distributional shift, resulting in instability. Figure \ref{fig:smallgap_offline} illustrates that ESR-O2O significantly outperforms other methods when provided with medium', medium-replay', and `replay' data, while still maintaining comparable performance with the top-performing baseline method, IQL. In certain tasks such as Walker, ESR-O2O even surpasses the performance of IQL.

\clearpage

\subsection{Results in the MetaWorld benchmark.}
\begin{figure}[htbp]
    \centering
    \includegraphics[width=0.5\linewidth]{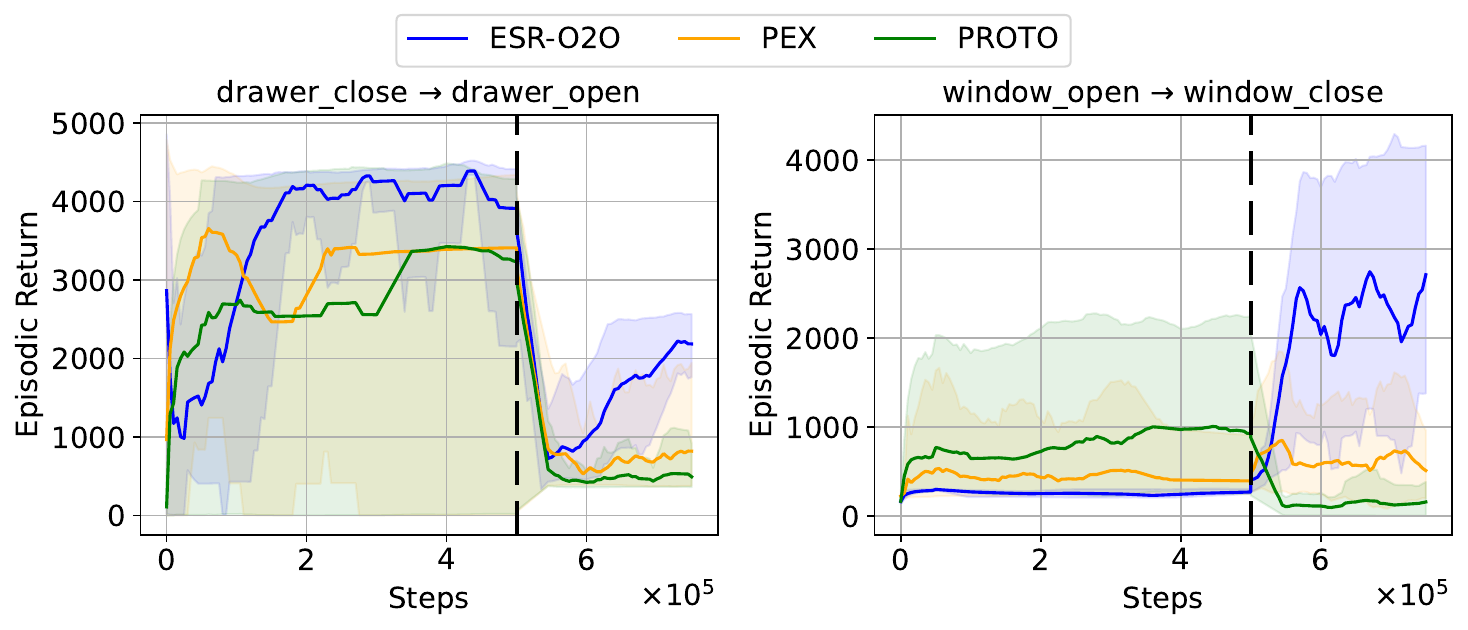}
    \caption{Fine-tuning performance in MetaWorld tasks. Since these manipulation tasks are more challenging than Mujoco locomotion tasks, current offline-to-online RL methods cannot perform well during fine-tuning like the results in Figure \ref{fig:smallgap}. In contrast, ESR-O2O is robust to reward gaps and successfully adapts to the new environment.}
    \label{fig:meta}
\end{figure}

We have conducted experiments on the MetaWorld benchmark, which contains a variety of challenging tasks and serves as an excellent platform for evaluating generalization capabilities in the presence of reward gaps. As shown in Table \ref{tab:tasks}, we consider four specific tasks, including the transition from 'drawer-close-v2' to 'drawer-open-v2' and the transition from 'window-open-v2' to 'window-close-v2'. Although the offline and online tasks share the same dynamics function, they have distinct reward functions. These manipulation tasks are notably more challenging compared to the DMC Mujoco locomotion tasks. In addition to the tabular comparison provided in Table \ref{tab:meta}, we also present average return curves across five random seeds in Figure \ref{fig:meta}. These results clearly demonstrate the superior performance of our method during fine-tuning when compared to PEX and PROTO, underscoring our method's exceptional generalization ability.

\section{More explanations about ablation studies.}

\paragraph{Whether using ensembles:} As illustrated in Figure \ref{fig:ablation}, eliminating the ensemble of SR significantly impairs the fine-tuning performance. In scenarios with a large reward gap, such as the reach task, we observe that useful policies cannot be learned without the ensemble of SR. This effect is even more pronounced when the available data is generated from narrow data distributions. For scenarios with a smaller reward gap, such as the quadruped task, the performance drop is relatively small when eliminating the ensemble of SR alone, but becomes substantial when both the ensemble of SR and Q networks are discarded. In summary, the ensemble of SR is crucial for addressing scenarios with a large reward gap. Additionally, the ensemble of Q networks plays a significant role when dealing with scenarios with a small reward gap or when the offline data is abundant.

\paragraph{The quantity of ensembles:} In our extended experiments, we systematically varied the ensemble's network count from 2 to 10, and the results are presented in Figure \ref{fig:ablation2} of the revised manuscript. Our results revealed that when we use more than 6 ensemble networks, the performance is consistently stable. Meanwhile, the stability and performance of offline-to-online adaptation improve as the number of networks increases. In contrast, reducing the number of networks to 4 led to a substantial decrease in performance in specific tasks. Notably, when the network count reduces to 2, the networks struggled to learn effective representations through online fine-tuning. Furthermore, it is worth noting that we implemented a network parallelization technique in our code, similar to the approach described in [1]. This parallelization method substantially mitigates the computation time overhead associated with increasing the number of networks.

\section{More experimental results in Reach, Quadruped, and Walker.}
\begin{figure}[ht]
    \centering
\includegraphics[width=0.9\linewidth]{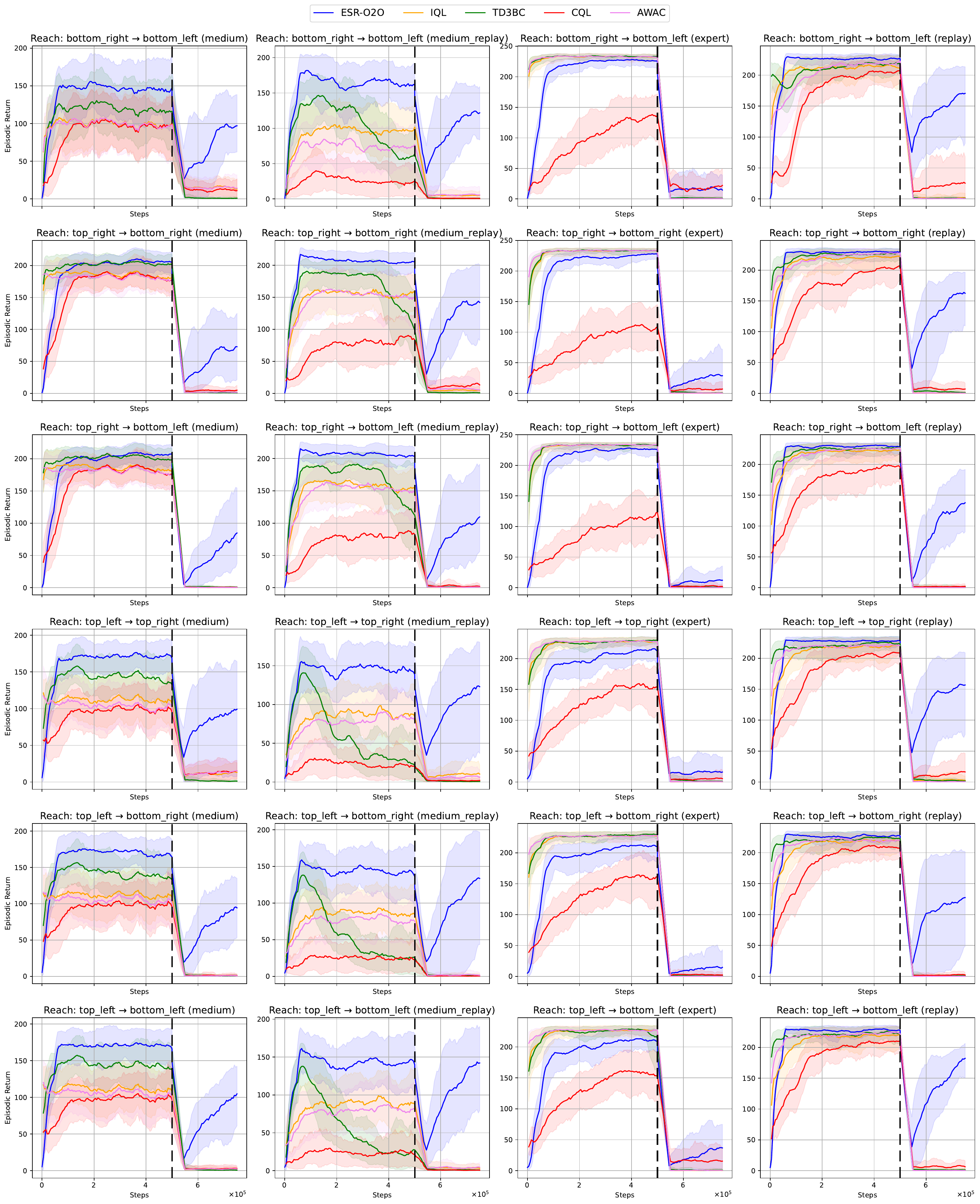}
    \caption{Performance comparisons on reach tasks.}
    \label{fig:append-reach}
\end{figure}

\begin{figure}[ht]
    \centering
\includegraphics[width=0.9\linewidth]{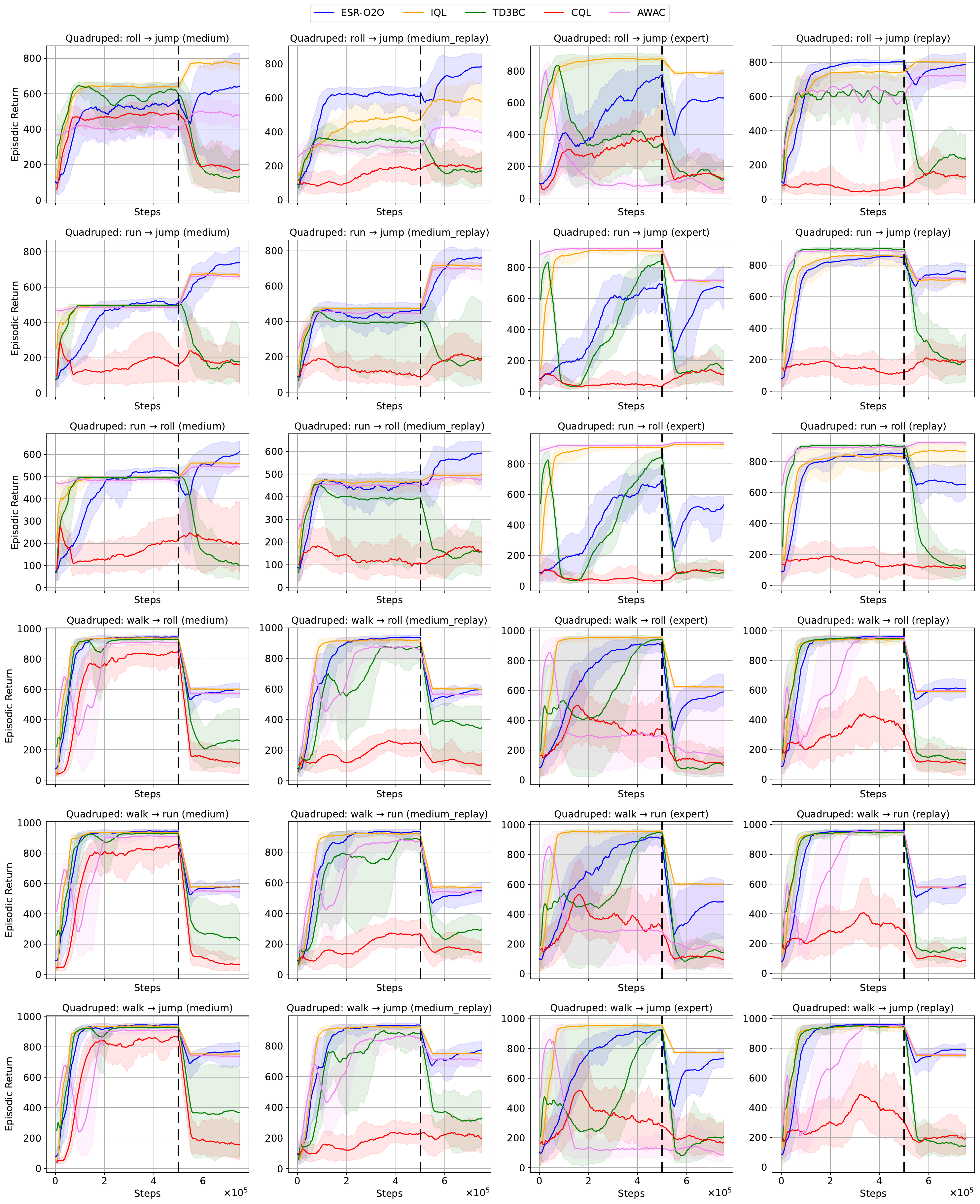}
    \caption{Performance comparisons on quadruped tasks.}
    \label{fig:append-quad}
\end{figure}

\begin{figure}[ht]
    \centering
\includegraphics[width=0.9\linewidth]{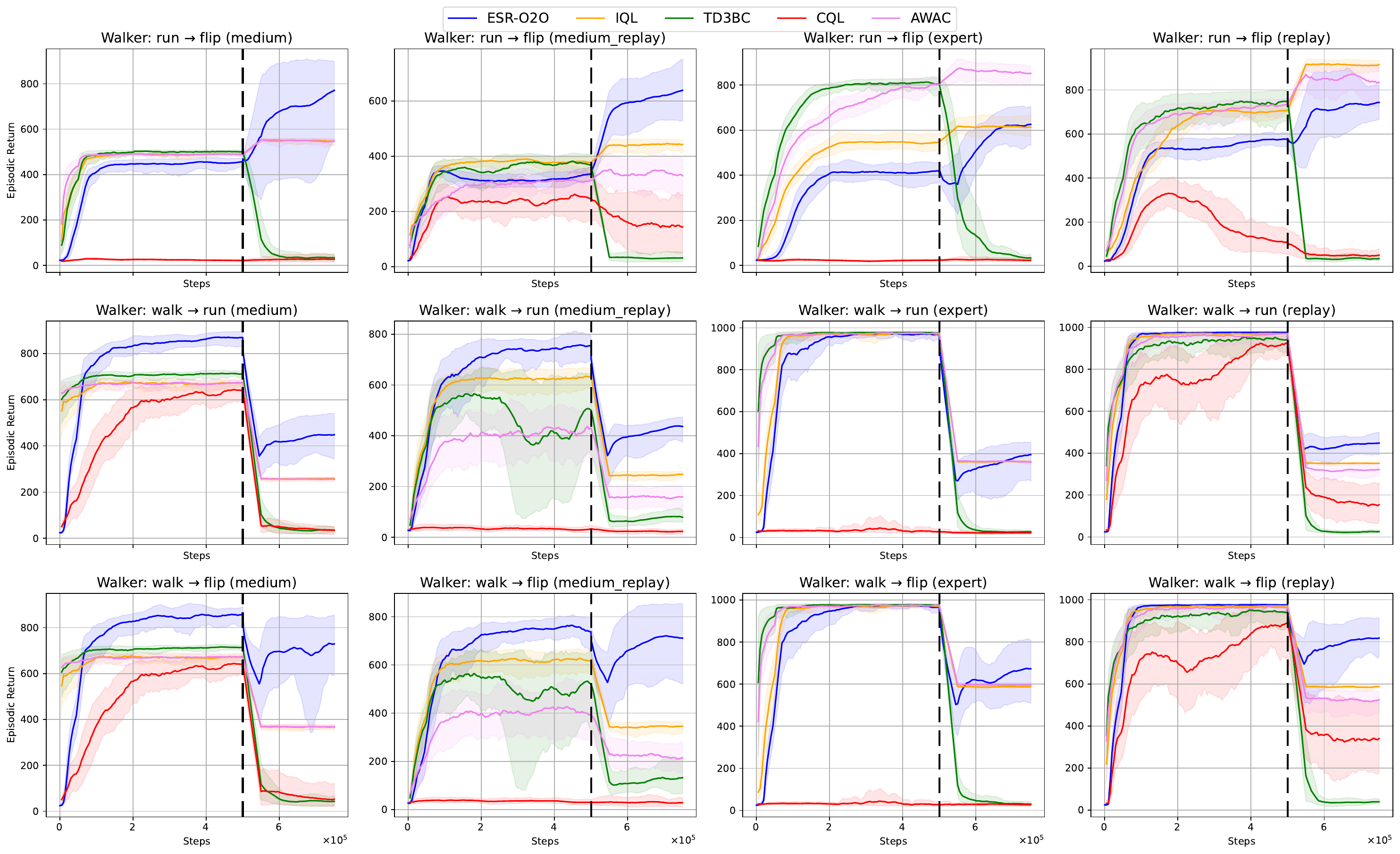}
    \caption{Performance comparisons on walker tasks.}
    \label{fig:append-walk}
\end{figure}

Additional experimental results about the fine-tuning performance are presented on Figures \ref{fig:append-reach}, \ref{fig:append-quad}, and \ref{fig:append-walk}.

\end{appendix}

\end{document}